\documentclass[11pt]{article}

\usepackage{geometry}
\usepackage{setspace}
\usepackage{caption} 
\usepackage{graphicx}
\usepackage{url}
\usepackage{array}
\usepackage{svg}
\usepackage{pdflscape}
\usepackage[acronym,nonumberlist,nopostdot,notree,nomain]{glossaries}

\newacronym{bpm}{BPM}{business process management}
\newacronym{dsr}{DSR}{design science research}
\newacronym{dnn}{DNN}{deep neural network}
\newacronym{lstm}{LSTM}{long short-term memory}
\newacronym{gru}{GRU}{gated recurrent unit}
\newacronym{gpt}{GPT}{generative pre-trained transformer}
\newacronym{cnn}{CNN}{convolutional neural network}
\newacronym{rnn}{RNN}{recurrent neural network}
\newacronym{bert}{BERT}{bidirectional encoder representations from transformers}
\newacronym{ml}{ML}{machine learning}
\newacronym{dl}{DL}{deep learning}
\newacronym{it}{IT}{information technology}
\newacronym{is}{IS}{information systems}
\newacronym{erp}{ERP}{enterprise resource planning}
\newacronym{ppm}{PPM}{predictive process monitoring}
\newacronym{llm}{LLM}{large language model}
\newacronym{tl}{TL}{transfer learning}
\newacronym{bce}{BCE loss}{binary cross entropy loss}
\newacronym{mcc}{MCC}{Matthews correlation coefficient}
\newacronym{xgb}{XGBoost}{extreme gradient boosting}
\newacronym{itil}{ITIL}{Information Technology Infrastructure Library}
\newacronym{iso}{ISO}{International Organization for Standardization}
\newacronym{iec}{IEC}{International Electrotechnical Commission}
\newacronym{gdpr}{GDPR}{General Data Protection Regulation}
\newacronym{csrd}{CSRD}{Corporate Sustainability Reporting Directive}
\newacronym{gaap}{GAAP}{Generally Accepted Accounting Principles}
\newacronym{ifrs}{IFRS}{International Financial Reporting Standards}
\newacronym{itsm}{ITSM}{information technology service management}
\newacronym{roberta}{RoBERTa}{robustly optimized BERT pretraining approach}
\newacronym{sbert}{SBERT}{sentence BERT}
\newacronym{ict}{ICT}{information and communications technologies}
\newacronym{mlp}{MLP}{multi-layer perceptron}
\newacronym{glove}{GloVe}{pre-trained global vectors for word representation}
\newacronym{adam}{Adam}{adaptive moment estimation}
\usepackage{mathptmx}
\usepackage[T1]{fontenc}
\usepackage{amsmath}
\usepackage{csquotes}
\usepackage{booktabs}
\usepackage{hyperref}
\usepackage{comment}
\usepackage{xcolor}
\usepackage{amssymb}
\DeclareMathAlphabet{\pazocal}{OMS}{zplm}{m}{n}
\SetMathAlphabet\pazocal{bold}{OMS}{zplm}{bx}{n}
\usepackage{multirow}

\usepackage{pifont}
\newcommand{\cmark}{\ding{51}}%
\newcommand{\xmark}{\ding{55}}%

\usepackage{natbib}
\setcitestyle{aysep={}}

\newenvironment{biseabstract}{%
\begin{quote} \bf}
{\end{quote}}

\newenvironment{bisekeywords}{%
\begin{quote} \it \textbf{Keywords:}}
{\end{quote}}

\title{From Source to Target: Leveraging Transfer Learning for Predictive Process Monitoring in Organizations} 

\author
{Sven Weinzierl$^{1\ast}$, Sandra Zilker$^{1, 2}$, Annina Liessmann$^{1}$, Martin Käppel$^{1}$, \\ Weixin Wang$^{1}$, and Martin Matzner$^{1}$\\
\\
\normalsize{$^{1}$Chair of Digital Industrial Service Systems, Friedrich-Alexander-Universität Erlangen-Nürnberg,}\\
\normalsize{Fürther Str. 248, 90429 Nürnberg, Germany}\\
\normalsize{$^{2}$Professorship for Business Analytics, Technische Hochschule Nürnberg Georg Simon Ohm}\\
\normalsize{Hohfederstr. 40, 90489 Nürnberg, Germany}\\
\\
\normalsize{$^\ast$Corresponding author: Sven Weinzierl, E-mail: sven.weinzierl@fau.de}
}

\date{}

\begin{document} 

\baselineskip18pt


\maketitle

\begin{biseabstract}
  Event logs reflect the behavior of business processes that are mapped in organizational information systems. Predictive process monitoring (PPM) transforms these data into value by creating process-related predictions that provide the insights required for proactive interventions at process runtime. Existing PPM techniques require sufficient amounts of event data or other relevant resources that might not be readily available, which prevents some organizations from utilizing PPM. The transfer learning-based PPM technique presented in this paper allows organizations without suitable event data or other relevant resources to implement PPM for effective decision support. 
  This technique is instantiated in both a real-life intra- and an inter-organizational use case, based on which numerical experiments are performed using event logs for IT service management processes. 
  %
  The results of the experiments suggest that knowledge of one business process can be transferred to a similar business process in the same or a different organization to enable effective PPM in the target context. 
  The proposed technique allows organizations to benefit from transfer learning in intra- and inter-organizational settings by transferring resources such as pre-trained models within and across organizational boundaries. 
 
\end{biseabstract}

\begin{bisekeywords}
Predictive process monitoring, Process mining, Transfer learning, Language model
\end{bisekeywords}

\section{Introduction}

Business processes are the backbone of organizational value creation~\citep{dumas2018fundamentals}. With ongoing digitalization, business processes are increasingly mapped in \gls{is}, such as \gls{erp} systems, and executed by process users and other stakeholders~\citep{beverungen2021seven}. Consequently, business processes leave behind large amounts of digital event data that reflect their behavior \citep{van_der_aalst_2016}. To transform these data into value, various business process analytics approaches have been developed~\citep{ZurMuehlen2009}. 

One of these approaches is \emph{\gls{ppm}} -- a set of techniques that leverage historical event log data \citep{maggi2014predictive} to create prediction models \citep{grigori2004business}, capable of predicting various targets in running business process instances, such as the next activity, remaining time, or process-related outcomes~\citep{di2018predictive}. 
By creating these process-related predictions, \gls{ppm} enables proactive interventions in running business process instances, helping to reduce risks and optimize resource allocation \citep{di2018predictive}. This leads to improved business process performance \citep[][]{marquez.2017}, even across organizational boundaries \citep{oberdorf2023predictive}. Therefore, \Gls{ppm} is gaining momentum in \gls{bpm}, while its techniques are considered a new class of process mining techniques~\citep{di2022predictive}. 

Most \gls{ppm} techniques rely on shallow \gls{ml} models \citep{marquez.2017} or \gls{dl} models \citep{rama2021deep} to generalize knowledge from historical to unseen event log data~\citep{weinzierl2024machine}.  
These models are trained using historical event log data and then applied to new instances of a running business process in order to predict various targets depending on the business context and the associated analytical goals~\citep{maggi2014predictive}.
However, organizations -- or different departments or subsidiaries within the same organization -- often lack the prerequisites for building effective prediction models. These include access to sufficient amounts of event log data \citep{Kaeppel2021} and the technical infrastructure required to train complex models, such as graphical processing units \citep{arpteg2018software, rama2021deep}. 

Consequently, some organizations cannot benefit from \gls{ppm}, while others may already have access to richer event log data or (pre-)trained prediction models for similar business processes. If these processes are sufficiently similar in terms of semantics (e.g., meaning of activity names) and structure (e.g., similar control flow), it becomes possible to overcome the aforementioned obstacles by transferring a prediction model from one business process to another instead of training a new model from scratch. This idea is supported by the fact that many organizations implement standardized or partially standardized business processes (e.g., \gls{itsm} processes), which are mapped in \gls{erp} systems such as SAP \gls{erp}, Oracle \gls{erp} Cloud, or Microsoft Dynamics 365. 

\begin{figure}[ht]
    \centering
    \includegraphics[width=1\linewidth]{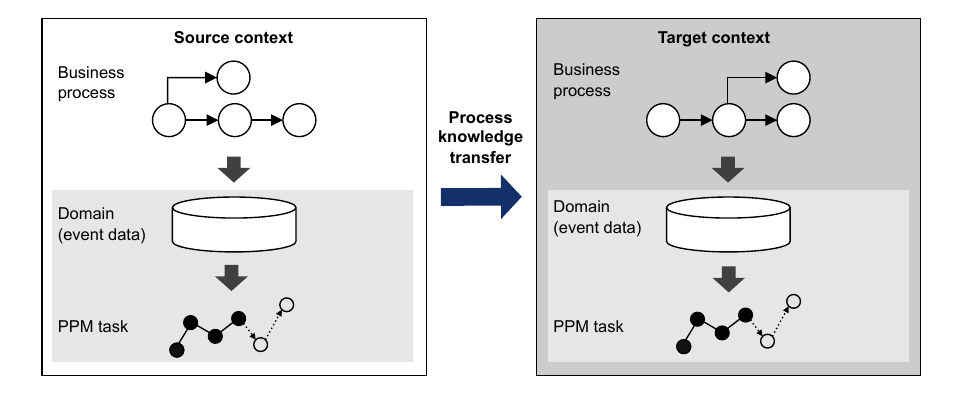}
    \caption{Transfer process knowledge from source to target context in \gls{ppm}.}
    \label{fig:intro}
\end{figure}

The field of \emph{\gls{tl}} provides techniques specifically designed to support such scenarios \citep{pan2009survey}, enabling the transfer of \gls{ml} models and the knowledge they contain from one learning problem (i.e., domain and task) to another. 
In \gls{ppm}, \gls{tl} can be applied in \emph{intra-organizational} scenarios, in which a prediction model is transferred from one process to another within a single organization (e.g., across departments or subsidiaries), and in \emph{inter-organizational} scenarios, where prediction models are transferred between processes of different organizations. 
In both scenarios, we consider all resources to be transferred as process knowledge. This includes the prediction model (also called the base model). Process knowledge originates in its source context, which includes the business process itself, the event data produced by it, and the specific \gls{ppm} task being addressed using the event data of the business process. The transferred process knowledge is then applied in the target context (see~Figure~\ref{fig:intro}).

Building on this idea, we present a \gls{tl}-based technique for \gls{ppm} that relies on existing similarities between business processes and aims to apply a prediction model to a target context. Therefore, the prediction model is transferred together with further relevant resources, such as information on the corresponding encoding strategy. In particular, this work contributes to the existing body of knowledge in four ways:  

\begin{enumerate}

    \item We present a \gls{tl}-based technique for \gls{ppm} that enables the transfer of a base model for an outcome-oriented prediction target without the necessity to fine-tune it on the event data from the target context. Our approach also provides guidance on which resources need to be transferred, including pre-trained models, information on the corresponding encoding strategy, further preprocessing details, and metrics used for measuring the prediction performance.

    \item We propose a cross-domain encoding strategy for event data that enables effective knowledge transfer from a source to a target context. While timestamp information is encoded via a relative mapping approach, activity information is encoded using pre-trained embedding models to preserve semantic consistency across contexts.
    
    \item We propose a two-layer \gls{lstm} model with dedicated parameter initialization as the prediction model of the source context, used for mapping the encoded traces onto process-related outcome predictions. This model improves learning from the event data of the source context and facilitates knowledge transfer to the target context.
    
    \item We empirically evaluate our \gls{tl}-based \gls{ppm} technique in an intra- as well as an inter-organizational \gls{itsm} use case. The results show that the transferred base model achieves a clearly higher prediction performance than models built with traditional \gls{ppm} techniques on the event data of the target context.  
\end{enumerate}

To carry out our research, we adopt the computational design science paradigm \citep{rai2017editor} and structure our work according to the \gls{dsr} publication schema by \citet{Gregor.2013}:  
First, we present relevant background on \gls{ml} and \gls{dl} in \gls{ppm}, and \gls{tl} in general (Section~\ref{sec:background}).
Second, we explain the practical relevance of \gls{tl} for \gls{ppm} in organizations and introduce inter- and intra-organizational \gls{tl} in the context of \gls{ppm} (Section~\ref{sec:PracticalRelevance}).  
Subsequently, we give an overview of related work on \gls{tl} in \gls{ppm} and position our proposed technique with regard to it (Section~\ref{sec:relatedwork}). 
We then provide a detailed description of the proposed artifact~(Section~\ref{sec:artifact}). 
Following this, the evaluation of our artifact is described by introducing the intra- and inter-organizational use case and providing details on the setup used in the evaluation~(Section~\ref{sec:evaluation}). 
Finally, we present the results of our evaluation (Section~\ref{sec:Results}), discuss implications and future research directions (Section~\ref{sec:discussion}), and conclude the paper~(Section~\ref{sec:conclusion}).

\section{Background}
\label{sec:background}
\subsection{Event log data and predictive process monitoring}
A \emph{business process} is defined as a sequence of interrelated activities and decisions undertaken to achieve a valuable outcome for a customer~\citep{dumas2018fundamentals}. Each execution of a business process is referred to as a \emph{process instance} or \emph{case}~\citep{van_der_aalst_2016}. \Gls{it} systems (e.g., \gls{erp} systems) commonly record these executions in the form of event logs, that is, collections of timestamped events described by various event attributes that store information about the execution of activities (i.e., steps in the process)~\citep[][p.~130]{van_der_aalst_2016}. Every event contains at least a case identifier, the name of the executed activity, and a timestamp indicating when the event occurred. Events belonging to the same process instance can be temporally ordered by their timestamp to form a so-called \emph{trace}~\citep[][p.~134]{van_der_aalst_2016}.

Event logs serve as the main input for \gls{ppm} techniques.
\Gls{ppm} provides a set of techniques that utilize event log data to make predictions on the future behavior or properties of ongoing process instances \citep{verenich2019survey}. These techniques aim to support proactive decision-making to enhance process performance and counteract potential risks during process execution \citep{marquez.2017}.
\Gls{ppm} techniques can target various types of predictions. One common prediction task is \emph{outcome prediction}, which aims to forecast whether a process instance will end with a certain (often categorical) outcome~\citep{Teinemaa.2019}. 
Other prediction targets include the next activity or the remaining sequence of activities (suffix), as well as temporal aspects such as the timestamp of the next activity or the remaining time until process completion~\citep{verenich2019survey}.

\subsection{Machine learning in predictive process monitoring}
\label{sec:background_MLinPPM}
A branch of early \gls{ppm} approaches \citep[e.g.,][]{vanderAalst.2011b,RoggeSolti.2015} create prediction models by augmenting given or discovered process models with predictive capabilities~\citep{oberdorf2023predictive}. These \gls{ppm} approaches are called \emph{process-aware} because they use prediction models that exploit an explicit representation of a process model (e.g., a probabilistic finite automaton or a stochastic Petri net) to make a prediction \citep{marquez.2017}. 
However, as real-world processes are usually more complex than reflected in process models~\citep{van2011processDis}, the process-model dependence limits the generalizability of the prediction models.  
To overcome this limitation, a more recent branch of \gls{ppm} approaches \citep[e.g.,][]{maggi2014predictive,tax2017predictive} proposes encoding an event log's traces as feature vectors for the straightforward building of models using \gls{ml} algorithms~\citep{oberdorf2023predictive}. These approaches are called \emph{non-process-aware} as the \gls{ml} models representing the prediction models do not use an explicit description of the process model~\citep{marquez.2017}. By leveraging the generalization capabilities of \gls{ml} models, the approaches often achieve a higher prediction performance than prediction models built on top of process models \citep{marquez.2017}. 

Besides a wide range of conventional \gls{ml} algorithms such as linear regressions, logistic regressions, decision trees, random forests, and gradient boosted decision trees, \gls{dl} algorithms in the form of \glspl{dnn} have been applied for building prediction models \citep{marquez.2017, Teinemaa.2019, verenich2019survey}.
%
They have gained prominence due to their ability to learn complex patterns and dependencies in large datasets~\citep{lecun2015deep}. Also in \gls{ppm}, techniques relying on \glspl{dnn} have shown superior prediction performance and generalizability across various \gls{ppm} tasks \citep{kratsch2021machine}. Therefore, \glspl{dnn} have become the state-of-the-art in \gls{ppm} for building prediction models. 

In recent years, a variety of neural network architectures have been employed for building models for different prediction targets, including \glspl{mlp}, \glspl{cnn}, and \glspl{rnn}, with either \gls{lstm} cells or \glspl{gru}~\citep{neu2022systematic,rama2021deep,weinzierl2024machine}. Among these, \glspl{rnn}, particularly \gls{lstm} networks, have been the most widely adopted in \gls{ppm}. 
For example, \citet{evermann2017predicting} apply \gls{lstm} models for the next activity prediction, whereas \citet{tax2017predictive} use \gls{lstm} models to predict both the next activity and its corresponding timestamp. 
%
More recently, transformer-based models, originally introduced in natural language processing, have shown high prediction performance. These models utilize self-attention mechanisms to effectively capture long-range dependencies in sequential data, as present in event data. For instance, \citet{bukhsh2021processtransformer} proposed a transformer model capable of predicting next activities, remaining time, and timestamps. 
Beyond efforts to enhance prediction performance through increasingly sophisticated and powerful neural network architectures, complementary strategies have focused on enriching the input data by incorporating additional sources of information. While many prediction models primarily rely on control-flow information and time information derived from timestamps (e.g., elapsed time), several approaches extend this by integrating contextual information, for example, in the form of case-level or resource attributes~\citep{weinzierl2019predictive}. For example, \citet{di2018predictive} propose multi-perspective models that combine control-flow and contextual information for outcome prediction. As another example, \citet{Hinkka.2019b} suggest clustering categorical event attributes and using the resulting cluster labels as additional information for a \gls{rnn}. In contrast to these approaches, \citet{Pegoraro2021TextAwarePM} and \citet{cabrera2022text} integrate unstructured data in the form of text as contextual information. 
The ability of \gls{dl} methods to handle diverse prediction tasks and to incorporate heterogeneous data sources not only demonstrates their flexibility but also hints at the potential for reusing learned representations in varying contexts~\citep{weinzierl2024machine}.

To build a prediction model using an \gls{ml} or \gls{dl} algorithm, it is crucial to appropriately encode the underlying event data, as the quality and structure of the encoding strongly affect the model’s ability to generalize. Thus, the encoding determines how transferable and reusable the learned representations are across different but related tasks. In many approaches, process instances are transformed into fixed-size feature vectors labeled with a prediction target, as required in supervised \gls{ml} settings, such as classification or regression~\citep{di2022predictive}. However, especially in \gls{dl}-based models, alternative representations such as event sequences or graph-structured inputs are also employed, allowing models to process richer structural information directly~\citep{Weinzierl.2022}.
Typically, two essential encoding steps are necessary: first, sequence encoding, to capture the sequential structure of the traces, and second, event encoding, to encode the information contained in the events (i.e., the event attributes). For the sequence encoding, various strategies have been proposed. These strategies range from simple boolean encodings, indicating the presence of specific activities, to more detailed representations, preserving the full control flow by including the complete sequence of events or the last $n$ events (states)~\citep{di2022predictive}. 
The encoding of the event attributes depends largely on their data types. Categorical attributes, such as activity names or resources, are either one-hot encoded or represented by an embedding vector, that is, a dense vector capturing the semantics. Numerical attributes usually require only normalization or scaling into a particular interval~\citep{Teinemaa.2019}. Temporal information, especially timestamp attributes, can be encoded by deriving features, such as \emph{time since the last event} or the \emph{weekday}, which then can be treated as numerical or categorical features~\citep{tax2017predictive}. 
Additional data attributes on the process context can be present in the form of text. For these attributes, simple natural language processing techniques can be used, such as bag-of-n-grams or latent Dirichlet allocation \citep{teinemaa2016predictive}, to extract semantically shallow features from text data. 
By using more advanced techniques that are based on (large) language models, such as contextualized word embedding models like \gls{bert} or \gls{roberta}, semantically rich features can be extracted from text data~\citep{cabrera2022text,liessmann2024predicting}.

\subsection{Transfer learning}
Typical \gls{ml}-based approaches, like the \gls{ml}-based \gls{ppm} techniques described in Section \ref{sec:background_MLinPPM}, require the availability of labeled data with the same distribution in training and test sets~\citep{pan2009survey}. This might not be the case in practice. 
\Gls{tl} offers a solution, as it \textquote{allows the domains, tasks, and distributions used in training and testing to be different} \citep[][p.~1346]{pan2009survey}. 
A domain $\pazocal{D} = \{\pazocal{X}, P(X)\}$ is described by the feature space $\pazocal{X}$, that is, the set of all term vectors, and the marginal probability distribution $P(X)$, where $X$ is an instance set $\left\{\mathbf{x} \mid \mathbf{x}_i \in \pazocal{X}, \text{with } i=1,\dots,n\right\}$. 
Task $\pazocal{T}$ consists of the label space $\pazocal{Y}$, that is, the set of all label values, and the predictive function $f(\cdot)$.
In \gls{tl}, a prediction model for the domain with limited data, that is, the target domain, can be improved by transferring knowledge from another domain, the source domain \citep{weiss2016survey}. 
More specifically, the learning of the target prediction function $f_T(\cdot)$, for the target learning task $\pazocal{T}_T$, and target domain $\pazocal{D}_T$, can be improved by transferring knowledge from the source domain $\pazocal{D}_S$, and source task $\pazocal{T}_S$, with either $\pazocal{D}_S \neq \pazocal{D}_T$ or $\pazocal{T}_S \neq \pazocal{T}_T$ \citep{pan2009survey}. 

In the context of \gls{ppm}, a domain is represented by the encoded event data $X$. The feature space $\pazocal{X}$ of the encoded event data $X$ is determined by the selected event attributes and the used sequence and event coding. The marginal probability distribution $P(X)$ of the encoded event data $X$ is the distribution of the features used to represent the event data $X$. The addressed \gls{ppm} task (e.g., the prediction of a process-related outcome) is denoted by $\pazocal{T}$. The prediction function $f(\cdot)$ of the task aims to map a running process instance to a label value $y \in \pazocal{Y}$, where $\pazocal{Y}$ is the space of all possible label values for the \gls{ppm} task~$\pazocal{T}$. 

 \begin{figure}[ht]
 \centering
 	\includegraphics[width=0.95\textwidth]{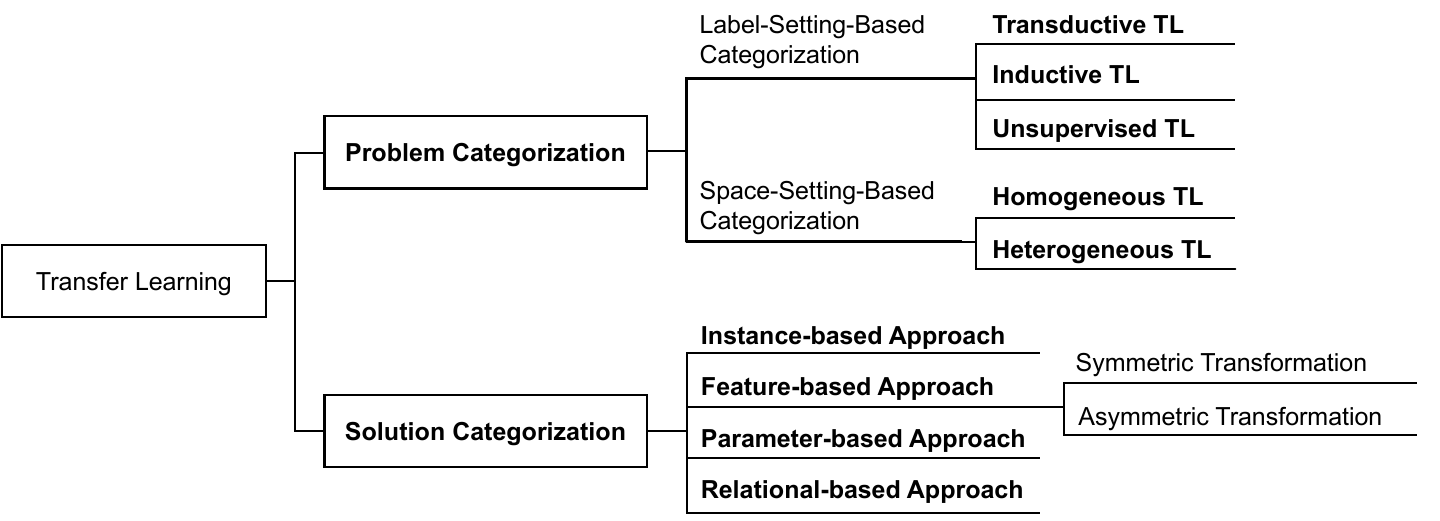}
 	\caption{Categories of \gls{tl} \citep{zhuang2020comprehensive}.}\label{fig:categoriesTL} 
 \end{figure}

The field can be categorized from a problem and a solution perspective, as shown in Figure~\ref{fig:categoriesTL} \citep{zhuang2020comprehensive}.
From the problem perspective, depending on the existence of the label, three categories exist: i) transductive \gls{tl} (label only present in $\pazocal{D}_S$, $\pazocal{D}_S \neq \pazocal{D}_T$, $\pazocal{T}_S = \pazocal{T}_T$), ii) inductive \gls{tl} (label present in $\pazocal{D}_T$, $\pazocal{T}_S \neq \pazocal{T}_T$), and iii) unsupervised \gls{tl} (label unknown in both domains, $\pazocal{T}_S \neq \pazocal{T}_T$) \citep{pan2009survey}. 
Another differentiation can be made by comparing the feature space of the source and the target domain. While homogeneous \gls{tl} refers to $\pazocal{X}^S = \pazocal{X}^T$, in heterogeneous \gls{tl}, the feature space of the source and the target domain is not the same ($\pazocal{X}^S \neq \pazocal{X}^T$) \citep{weiss2016survey}.
From a solution perspective, \citet{pan2009survey} suggest four categories of approaches based on what is transferred: i) instance-based, ii) feature-based, iii) parameter-based, and iv) relational-based. 
In \emph{instance-based} approaches, instances from the source domain are re-weighted to be used in the target domain. 
\emph{Feature-based} approaches aim to transfer knowledge with a learned feature representation. This can be done \emph{asymmetrically}, meaning transforming source features to align with target features, or \emph{symmetrically}, which means discovering a common new feature space~\citep{zhuang2020comprehensive}. 
In \emph{parameter-based} approaches, shared (hyper-)parameters or hyperpriors between the source and the target domain are transferred. 
Lastly, \emph{relational-based} approaches transfer knowledge in the form of relations between the source and target data, assuming that the data are not independent and identically distributed~\citep{pan2009survey}.

Since we aim to transfer process knowledge within and across organizational boundaries, data privacy issues might prohibit instance-based approaches. 
Assuming that the organizations concerned are operating independently and not identically, relational-based approaches might not be appropriate for our setting. 
However, feature-based and parameter-based approaches seem promising to transfer process knowledge within and across organizational boundaries.

\section{Practical relevance of transfer learning for predictive process monitoring in organizations}
\label{sec:PracticalRelevance}
Transferring and utilizing knowledge from a source context \textit{S} to a target context \textit{T} is particularly interesting in connection with business processes.
In organizations, \gls{it} systems are in place to support the execution of business processes to create value. The widespread use of \gls{erp} systems offering similar modules for purchasing, sales and distribution, production, warehouse and inventory management, human resource management, or financial and cost accounting makes business processes comparable across organizations. 
Alongside \gls{erp} systems, the use of function-specific software, such as Salesforce for customer relationship management or ServiceNow for \gls{it} service management, leads to a standardization of processes.
In addition to \gls{it} systems that offer a common ground for executing business processes, organizations can follow industry best practices and standards. For example, they can follow those of the \gls{iso}\footnote{\url{https://www.iso.org} (Accessed 23 July 2025)} or the \gls{iec}\footnote{\url{https://www.iec.ch/homepage} (Accessed 23 July 2025)}. In \gls{it}, standards include \gls{iso}/\gls{iec}~27001 for information security management systems or the \gls{itil}\footnote{\url{https://www.axelos.com/certifications/itil-service-management/} (Accessed 23 July 2025)} for \gls{it} service management. In manufacturing, \gls{iso}~9001 for quality management or Six Sigma for process improvement are common.
Besides the voluntary adherence to industry standards, the activities and processes of an organization are also governed by rules and regulations. Prime examples of such regulations and directives are the \gls{gdpr}\footnote{\url{https://gdpr.eu} (Accessed 23 July 2025)} and the recently introduced \gls{csrd}\footnote{\url{https://finance.ec.europa.eu} (Accessed 23 July 2025)}. Similarly, organizations must follow certain accounting standards according to their location, such as the \gls{gaap}\footnote{\url{https://fasab.gov/accounting-standards/} (Accessed 23 July 2025)} or the \gls{ifrs}\footnote{\url{https://www.ifrs.org} (Accessed 23 July 2025)}.
Lastly, organizations have similar objectives based on which their processes are designed. For example, an organization can strive for cost or time reduction, improved efficiency, increased product or service quality, or customer satisfaction in their process executions.
The use of comparable software tools, adherence to regulations and (industry) standards, and common objectives make different business processes within and across organizations increasingly similar to each other.

 \begin{figure}[htbp]
 \centering
 	\includegraphics[width=\textwidth]{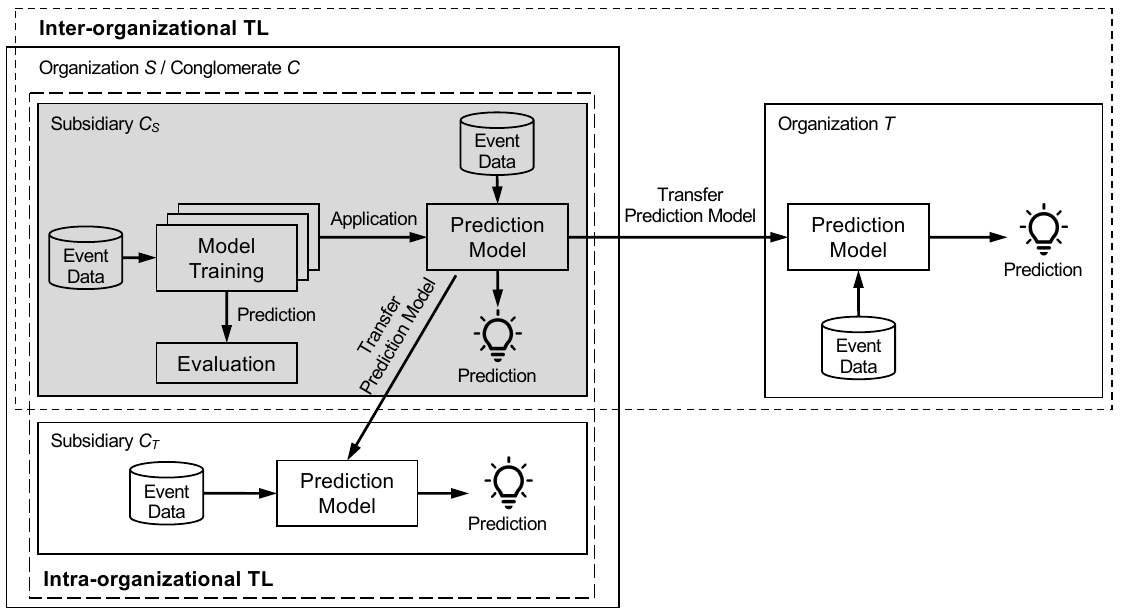}
        \caption{Intra- and inter-organizational \gls{tl} for \gls{ppm}.\protect\footnotemark}
    \label{fig:inter-intra-org}
 \end{figure}
\footnotetext{For simplification, Figure~\ref{fig:inter-intra-org} shows the transfer of the prediction model representative for all transferred resources from the source to the target context.}

This offers opportunities to transfer process knowledge within one organization (see~Figure~\ref{fig:inter-intra-org}). \textit{Intra-organizational \gls{tl} for \gls{ppm}} allows an organization to transfer knowledge across its processes.  
This might be especially useful when considering an organization operating with multiple subsidiaries in various locations. Such conglomerates often maintain multiple \gls{erp} systems for their subsidiaries based on industry or location. For example, an organization \textit{C} has \gls{ppm} with event data from its \gls{erp} system already in use in a source subsidiary $C_S$. To harness existing process knowledge in subsidiary $C_S$, a target subsidiary $C_T$ can utilize subsidiary $C_S$'s prediction models for its process data.
This becomes inherently convenient when laws and regulations inhibit the collection of sufficient historical process data for subsidiary $C_T$ to train its own prediction models.
In addition, intra-organizational \gls{tl} for \gls{ppm} offers the possibility for an organization to go beyond \gls{ppm} pilot projects and provide process owners with the tools to implement \gls{ppm} by themselves. Pre-trained prediction models for certain process types can be accessed by process owners to apply \gls{ppm} without the need for costly resources and extensive model training, almost like an internal model-as-a-service offering.

While transferring process knowledge through \gls{tl} within one organization mainly benefits the same organization, \gls{tl} for \gls{ppm} can also be applied across organizational boundaries (see~Figure~\ref{fig:inter-intra-org}). 
We coin this \emph{inter-organizational \gls{tl} for \gls{ppm}}.
For example, a target organization \textit{T} is interested in the application of \gls{ppm} and plans to launch a pilot project to investigate its use for one of its support processes, incident management. Organization \textit{T} logs all information related to solving incidents in its service desk using ServiceNow, following \gls{itil} while adhering to \gls{iso}/\gls{iec}~27001. With \gls{ppm}, organization \textit{T} aims to reduce the resolution time. To keep resources and costs low and to be able to launch a proof-of-concept relatively fast, organization \textit{T} can use a transferred prediction model, trained with event data of a similar incident management process of source organization \textit{S} instead of training a prediction model from scratch. 
The acquisition of organization \textit{T} by organization \textit{S} represents another use case for inter-organizational \gls{tl} for \gls{ppm}. Some processes of organization \textit{S} are already supported by \gls{ppm}. As part of the merger, organization \textit{S} wants to integrate organization \textit{T} and its \gls{erp} system into its \gls{ppm} efforts. To enable smooth integration, the prediction models used for processes in organization \textit{S} can be transferred to similar processes in organization~\textit{T}.

The increasing similarities of business processes create an application area for intra- and inter-organizational \gls{tl} in \gls{ppm}.
As inter-organizational \gls{tl} for \gls{ppm} aims to transfer process knowledge across organizations, the domains (i.e., the feature space and marginal probability distribution of event data) of the source and target contexts tend to differ more than in intra-organizational \gls{tl} for \gls{ppm}, in which knowledge is transferred between business processes within the same organization.
This greater difference between the source and target domains in an inter-organizational setting can be attributed to, for example, organization-specific standards and regulations for performing the same type of business process.  

\section{Related work}
\label{sec:relatedwork}
Related \gls{tl}-based \gls{ppm} techniques can generally be categorized into two groups: i) \gls{ppm} techniques that transfer resources within one business process in an organization (i.e., $\pazocal{D}_S$ and $\pazocal{D}_T$ represent event data on the same business process), and ii) \gls{ppm} techniques that transfer resources between similar business processes within one organization or across different organizations (i.e., $\pazocal{D}_S$ and $\pazocal{D}_T$ represent event data on similar business processes within the same or different organizations).

Some \gls{ppm} techniques in the first group aim to transfer parameters between models in transition system-based approaches, to capture patterns in event data better, and ultimately improve prediction performance.  
\citet{cao2022transition} present an approach for remaining time prediction. This approach combines Petri net-based process modeling and \gls{dl}. A Petri net is mined from event log data to construct a transition sequence, and prefixes are classified into partitions (transition division) based on their last activity. Autoencoder models are then trained on the prefixes within each partition to reduce dimensionality. Based on the prefixes encoded via the corresponding autoencoder model, an \gls{mlp} model is trained for each partition. \Gls{tl} is utilized by transferring parameters from the \gls{mlp} model trained on the event data of the previous partition ($\pazocal{D}_S$) to the \gls{mlp} model of the current partition, and fine-tuning the \gls{mlp} model of the current partition with available event data ($\pazocal{D}_T$). 
In a similar approach, \citet{ni2022predicting} introduce an autoencoder-based transition system for predicting the remaining time of a process. They mine a transition system from event log data and use a sparse autoencoder to compactly represent the event log traces. For each state of the transition system, an \gls{mlp} is pre-trained on the encoded event data of all states ($\pazocal{D}_S$) to capture global dependencies. These models' parameters are then transferred to new models, which are fine-tuned using state-specific event data~($\pazocal{D}_T$). 

Other \gls{ppm} techniques in the first group aim to transfer parameters between \gls{dl} models to improve prediction performance.   
\citet{mehdiyev2020novel} propose a technique for the next activity prediction, in which two autoencoder models are trained in a semi-supervised manner on available event data ($\pazocal{D}_S$). The parameters from the hidden layers of these models are then transferred to a prediction model for the next activity prediction, which is subsequently fine-tuned using the same event data ($\pazocal{D}_T$). 
The work of \citet{folino2019learning} suggests a technique for outcome prediction with limited labeled event data. An \gls{lstm}-encoder model is first trained on event data ($\pazocal{D}_S$) for the next activity prediction, as no extra labels are required for this target. The last hidden state of the encoder model's final \gls{lstm} layer is then transferred to a second model for outcome prediction. This model is then fine-tuned on a portion of the same event data, for which a process-related outcome label is available ($\pazocal{D}_T$).

Besides parameter-based \gls{tl}, \citet{cabrera2022text} propose a \gls{ppm} technique that leverages feature-based \gls{tl}. This is achieved by using various \gls{bert} models -- pre-trained on a huge corpus of text data ($\pazocal{D}_S$) -- to encode the text information available in event data ($\pazocal{D}_T$) and adding the encoded information to the input of a multi-tasking \gls{lstm} model. This \gls{lstm} model addresses the next activity prediction and the next timestamp prediction. 
In addition, the work suggests that fine-tuning the pre-trained \gls{bert} models on available event data using the next activity prediction task enhances the \gls{lstm} model's performance for predicting next activities and timestamps as it adapts the representation of the \gls{bert} models to the process control-flow.    

Some \gls{ppm} techniques in the first group also perform both parameter- and feature-based \gls{tl}.
\citet{pfeiffer2021multivariate} propose a \gls{ppm} technique combining Gramian angular fields and representation learning. The Gramian angular fields are applied to encode the event data as 2D images, while representation learning is performed to obtain a generic process representation from the encoded event data ($\pazocal{D}_S$). 
This representation is obtained using a two-step approach. First, a \gls{cnn} model is trained to map the encoded input onto trace variants. Second, the pre-trained \gls{cnn} model is extended with a fully-connected layer and heads for various prediction tasks, such as the next activity and next timestamp prediction, and then fine-tuned on the encoded event data. 
Subsequently, all heads except the desired one are removed, and the model is fine-tuned again on the encoded input ($\pazocal{D}_T$). By learning a model capturing a generic process representation and aligning it to various prediction tasks, feature- and parameter-based \gls{tl} is performed within one model.  
In a similar approach, \citet{chen2022multi} propose to learn a generic representation of traces by pre-training a \gls{bert} model on a set of unlabeled trace data ($\pazocal{D}_S$) using a custom masked activity modeling task. The pre-trained \gls{bert} model is then fine-tuned on another set of labeled trace data ($\pazocal{D}_S$) for the next activity or outcome prediction. 
As in the approach of \citet{pfeiffer2021multivariate}, one model capturing a generic process representation is learned and aligned to prediction tasks. Therefore, this approach also performs feature- and parameter-based \gls{tl}. 
\citet{pasquadibisceglie2023darwin} suggest using \gls{tl} for handling concept drifts in \gls{ppm}. Specifically, once a drift is detected within event data ($\pazocal{D}_S$), the parameters of both an \gls{lstm} model for the next activity prediction and a word2vec model for activity encoding are updated with current event data ($\pazocal{D}_T$). This enhances the prediction performance of the \gls{lstm} model and enables the word2vec model to deal with newly occurring activities. 
However, even though these \gls{ppm} techniques incorporate parameter- and feature-based \gls{tl}, they are not designed to work with different business processes, where the values and distributions of event attributes in the source and target domains are generally different. Consequently, these \gls{ppm} techniques cannot be used in an intra- or inter-organizational \gls{tl} setting. 

The second group of \gls{ppm} techniques aims to transfer resources from one business process to another. 
\citet{luijken2023an} use event data $\pazocal{D}_S$ of a business process in the source context to pre-train an \gls{lstm} model and a \gls{gpt} model for predicting activity and timestamp suffixes of a prefix.
The parameters of these pre-trained models are then transferred to new models in the target context that address the same prediction tasks. 
After this transfer, some layers remain unchanged and others are fine-tuned using the event data $\pazocal{D}_T$ of a business process in the target context.
Although the authors demonstrate the feasibility of heterogeneous \gls{tl} with real-life event data by transferring model parameters from source to target contexts, their technique does not consider the embedding of different sets of activities in event data $\pazocal{D}_S$ and $\pazocal{D}_T$.
However, feature-based \gls{tl} is -- besides parameter-based \gls{tl} -- necessary to transfer process knowledge between similar business processes within the same or across different organizations, as the values and distributions of event attributes in the event data of the business processes can differ from each other.

\begin{landscape}
\begin{table}[ht]
\footnotesize
\centering
\caption{Overview of existing works applying \gls{tl} for \gls{ppm}.}
\label{tab:relatedwork}
\resizebox{1.4\textwidth}{!}{%
\begin{tabular}{@{}p{3cm}p{2cm}p{2cm}p{3cm}p{1.5cm}p{10cm}p{1.5cm}p{3cm}p{6cm}p{1.5cm}p{1.5cm}p{2.5cm}@{}}
\toprule
\textbf{Authors} & \textbf{Problem categorization} & \textbf{Solution categorization} & \textbf{Prediction model} & \textbf{Input data} & \textbf{Transfer goal} & \textbf{Different business processes} & \textbf{Activity encoding} & \textbf{Timestamp encoding} & \textbf{Fine-tuning in target not necessary} & \textbf{Prediction target} & \textbf{Data set} \\ \midrule

 \citet{folino2019learning} & Inductive, homogeneous & Parameter-based & LSTM-encoder model & ACT, RES, DA & Transfer process knowledge from a prediction model tailored to an auxiliary task, for which no labeled event data is required, to an prediction model tailored to a target prediction task, for which labeled event data is required, but scarcely available & \xmark & One-hot encoding & Extraction of six time-related features (hour, weekday, month, and times elapsed since the process instance’s start, previous event, and midnight), min-max normalization & \xmark & OP, NAP & BPIC12 \\
 
 \citet{pfeiffer2021multivariate} & Inductive, homogeneous & Parameter-based, feature-based & CNN model with task-specific layer & ACT, TS, RES, DA & Transfer process knowledge from a model capturing a generic business process representation for various single prediction tasks to prediction models for single tasks, to address them more effectively & \xmark & Integer encoding, min-max scaling, Gramian angular field transformation & Extraction of one time-related feature (duration since start), min-max normalization, Gramian angular field transformation & \xmark & RTP, NAP, OP & Helpdesk, BPIC12, BPIC13, BPIC17, BPIC20, MobIS \\

\citet{chen2022multi} & Inductive, homogeneous & Parameter-based, feature-based & Multi-tasking, BERT-based transformer model & ACT & Transfer process knowledge from a generic business process representation for multiple tasks to prediction models for single tasks, to address them more effectively & \xmark & Embedding layer, positional encoding & - & \xmark & NAP, OP & Helpdesk, BPIC12, Sepsis \\\\
 
\citet{ni2022predicting} & Transductive, homogeneous & Parameter-based & MLP model & ACT, TS, RES, DA & Transfer process knowledge from a generic prediction model pre-trained on event data (all states in a transition system) to a specific prediction model fine-tuned on event data (one state), to address a prediction task more effectively & \xmark & One-hot encoding, autoencoder & Extraction of four time-related features (year, month, week, and day), discretization, autoencoder & \xmark & RTP & Helpdesk, Hospital Billing, BPIC12, BPIC13 \\

\citet{cao2022transition} & Transductive, homogeneous & Parameter-based & MLP model & ACT, TS & Transfer process knowledge from a prediction model pre-trained on event data (previous transition division) to another prediction model fine-tuned on event data (current transition division), to address a prediction task more effectively & \xmark & One-hot encoding, autoencoder & Extraction of eight time-related features (e.g., month, execution time, and elapsed time), min-max normalization, autoencoder & \xmark & RTP & Four synthetic event logs, BPIC12, BPIC17, \\

 \citet{pasquadibisceglie2023darwin} & Transductive, heterogeneous & Parameter-based, feature-based & LSTM model & ACT, TS, RES & Transfer process knowledge from a prediction/embedding model pre-trained on real-life event data to a new prediction/embedding model fine-tuned on real-life event data, to handle concept drifts and address a prediction task more effectively & \xmark & Embedding model & - & \xmark & NAP & 12 BPIC event logs \\
 
\citet{mehdiyev2020novel} & Transductive, homogeneous & Parameter-based & MLP model & ACT, RES, DA & Transfer process knowledge from an embedding model pre-trained on real event data to a new prediction model fine-tuned on real event data, to address a prediction task more effectively & \xmark & Hash encoding & - & \xmark & NAP & BPIC12, BPIC13, Helpdesk \\
 
 \citet{cabrera2022text} & Inductive, heterogeneous & Feature-based & Multi-tasking LSTM model & ACT, TS, text & Transfer process knowledge from a generic embedding model pre-trained on a large corpus of text data to an embedding model fine-tuned on real-life event data, to address prediction tasks with a successive prediction model more effectively & \xmark & Embedding model & Extraction of three time-related features (time passed since previous event, time within the day, and time within the week), min-max normalization & \xmark & NAP, NTP & BPIC16 \\

\citet{rizk2023case} & Transductive, homogeneous & Parameter-based, feature-based & LSTM model & ACT & Transfer process knowledge from a prediction model pre-trained on synthetic event data to a prediction model fine-tuned on real-life event data, to address a prediction task more effectively & \cmark & One-hot encoding, embedding layer & - & \xmark & NAP & Synthetic loan process log, BPIC15, BPIC18 \\
 
\citet{luijken2023an} & Transductive, homogeneous & Parameter-based & Multi-tasking GPT/LSTM model with layer freezing & ACT, TS & Use process knowledge from a prediction model pre-trained on real-life event data to another prediction model fine-tuned on real-life event data, to address a prediction task more effectively & \cmark & One-hot encoding, embedding layer & Scalar timestamp, min-max normalization, embedding layer & \xmark & ASP, TSP & Helpdesk, RTFM, BPIC12, BPIC13, BPIC15, BPIC17, Sepsis \\
 
 \citet{nur2024predictive} & Transductive, heterogeneous & Parameter-based, feature-based & Transformer model with time-related feature block & ACT, TS & Transfer process knowledge from a prediction model pre-trained on real-life event data to another prediction model fine-tuned on real-life event data, to address a prediction task more effectively & \cmark & Embedding layer, positional encoding & Extraction of two time-related features (elapsed time and lagged time) & \xmark & RTP & Helpdesk, BPIC20, BPIC12, BPIC11 \\

\midrule 

\textbf{Ours} & Transductive, heterogeneous & Feature-based, parameter-based & LSTM-encoder model with dedicated parameter initialization & ACT, TS & Transfer process knowledge from a source context to a target context to enable effective PPM between similar processes within an organization or across different organizations & \cmark & Embedding model & Extraction of one time-related feature (duration since start), relative scaling & \cmark & OP & Helpdesk, BPIC14 
\\ 
\bottomrule
\multicolumn{9}{l}{Note. ACT = Activity, TS = Timestamp, RES = Resource, DA = Data attribute, OP = Outcome prediction, NAP = Next activity prediction, RTP = Remaining time prediction, ASP = Activity suffix prediction, TSP = Timestamp suffix prediction, BPIC = Business process intelligence challenge.}
\end{tabular}%
}
\end{table}
\end{landscape}

Furthermore, two \gls{ppm} techniques combine feature-based and parameter-based \gls{tl}.
\citet{rizk2023case} pre-train an \gls{lstm} model on synthetic event data ($\pazocal{D}_S$), transfer parameters of this model to a new one, and fine-tune the new model using other synthetic event data ($\pazocal{D}_T$) to solve the same prediction task (i.e., next activity prediction). 
The first layer of the \gls{lstm} model is an embedding layer, which enables matching the feature space of $\pazocal{D}_S$ to that of $\pazocal{D}_T$. The same experiment is performed for the event data of two different real-life event logs. 
\citet{nur2024predictive} suggest a technique with a transformer-based \gls{dl} model for remaining time prediction. The transformer-based \gls{dl} model consists of three components: a trace and positional encoding block, a transferable block (leveraging multi-head self-attention to capture semantic and temporal relationships), and a time-related feature block.
\gls{tl} is used in the technique to address the challenge of limited event data. This involves training the transformer-based \gls{dl} model (base model) on a large amount of event data ($\pazocal{D}_S$), transferring the trained parameters of its transferable block to a new model, and fine-tuning the new model on a smaller amount of event data ($\pazocal{D}_T$) to improve its prediction performance for remaining process times.  
Although both techniques incorporate the ideas of feature-based and parameter-based \gls{tl} in their models, they require a fine-tuning step on event data in the target context. However, this is difficult in intra- or inter-organizational \gls{tl} settings, where event data are not or very limitedly available in the target context.

Table~\ref{tab:relatedwork} summarizes the works identified, which use \gls{tl} for \gls{ppm}, and shows how they differ from our proposed technique in terms of dimensions, such as \emph{transfer goal}, \emph{different business processes}, and \emph{fine-tuning in target not necessary}.


\section{Transfer learning-based technique for predictive process monitoring}
\label{sec:artifact}

\begin{figure}[ht] 
\centering
	\includegraphics[width=\textwidth]{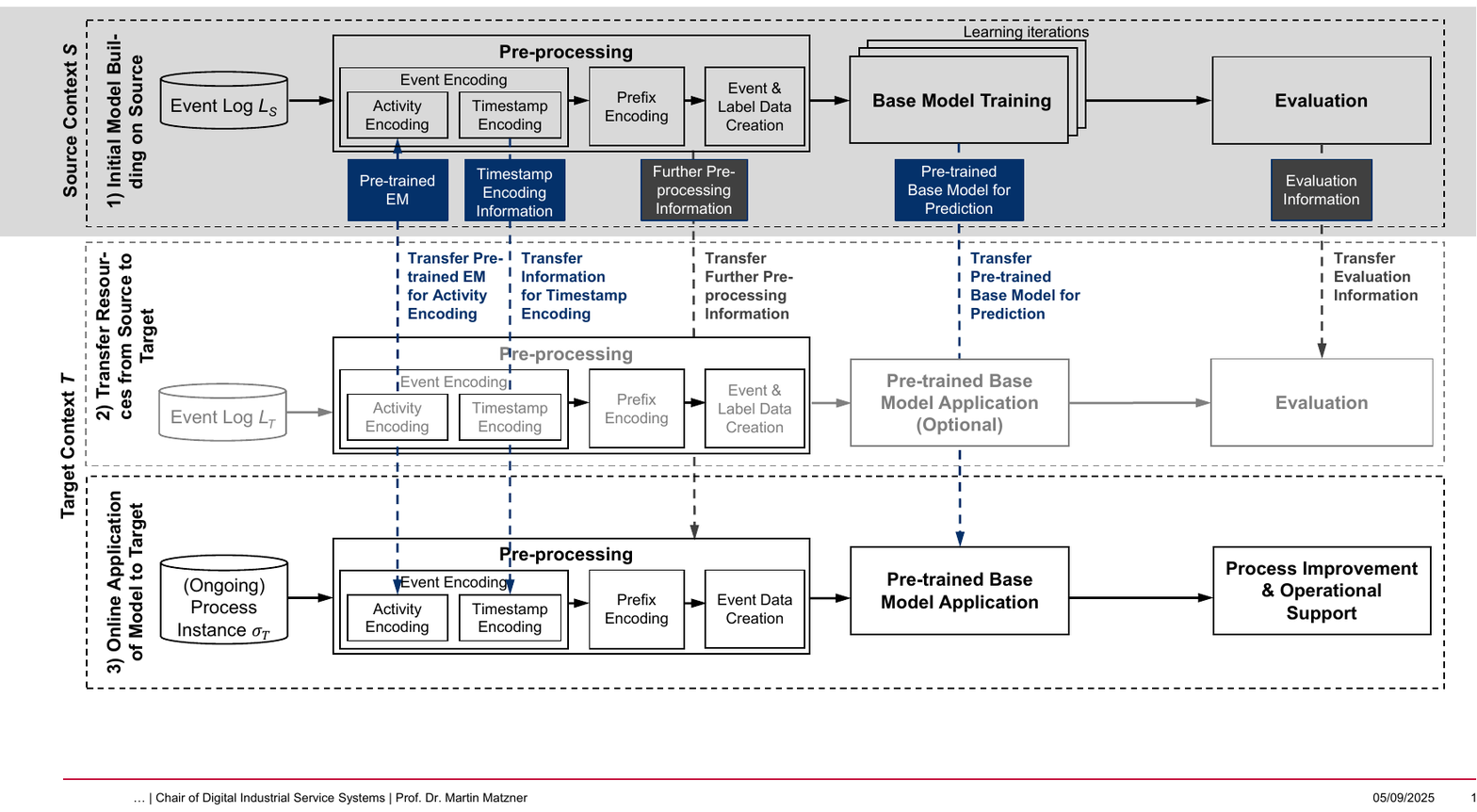}
	\caption{Overview of our \gls{tl}-based \gls{ppm} technique. \gls{tl}-specific and general resources that need to be transferred from the source context $S$ to the target context $T$ are colored blue and gray, respectively. \protect\footnotemark}
 \label{fig:artifact} 
\end{figure}

This section describes the design of our proposed \gls{tl}-based technique for \gls{ppm}. The technique is structured into three phases, as shown in Figure~\ref{fig:artifact}.    
In the first phase, \textit{initial model building on source}, an event log from the source context is loaded, preprocessed, and a \gls{dnn}-based base model is built and evaluated on preprocessed event data.
In the second phase, \textit{transfer resources from source to target}, relevant resources from the source context (e.g., pre-trained models) are transferred to the target context. If an event log from the target context is available, the base model from the source context can be applied and evaluated in an intermediate step before being used for (ongoing) process instances in the target context. 
In the third phase, \textit{online application of model to target}, an ongoing process instance of the target context is received. This process instance is then pre-processed using the transferred resources (e.g., information for timestamp encoding), and the transferred base model is applied to the pre-processed data. Finally, operational support is provided based on the base model's predictions for process improvement. 
In the following sections, the three phases of our proposed \gls{tl}-based technique for \gls{ppm} are described.

\footnotetext{The abbreviation \textquote{EM} in Figure~\ref{fig:artifact} stands for \textquote{embedding model}.}

\subsection{Initial model building on source}

The first phase, which aims to build and evaluate an initial base model $f_{\text{DNN}}(\cdot)$ in the source context $S$, consists of six steps.

First, an event log $L_S$ is loaded, representing a set of completed traces of a business process from the source context $S$. A complete trace $\sigma$ represents the data for one finished instance of a process, and contains a sequence of events, $e_1, \dots, e_N$, with sequence length $N$. 
The traces of the event log are assumed to be complete as they enable addressing \gls{ppm} tasks that refer to the end of a process instance (e.g., process-related outcome prediction).  
An event $e$ is a tuple $(p, a, t)$, where $p$ is the id of the process instance, $a$ is the process activity, and $t$ is the timestamp. In addition, to retain the temporal order of traces of the event log $L_S$, traces are sorted according to the timestamp of their first event. 

Second, the activities and the timestamps of the events of the event log $L_S$ are encoded. The activities are encoded by transforming their textual values into high-dimensional feature vector representations using a pre-trained embedding model $f_{\text{EMB}}: \sum^* \rightarrow \mathrm{R}^n$, where $\sum$ is the alphabet, $n$ is the embedding size, and $\sum^*$ are all finite activities that can be built from the alphabet $\sum$.
For example, word2vec~\citep{mikolov2013efficient} and \gls{bert}~\citep{kenton2019bert} are common natural-language-processing methods to create pre-trained embedding models. 
%
By encoding activity names in this way, they are represented more abstractly than, for example, in the case of a one-hot encoding, facilitating the transfer from a source to a target domain -- that is, symmetrical, feature-based \gls{tl} \citep{zhuang2020comprehensive}. Depending on the architecture and capability of the pre-trained embedding model used, the feature vectors representing the values of the activity attribute have different sizes. 
For example, a word2vec model is a small neural network model with one hidden layer, which learns the relationship between words and produces feature vectors of lengths, such as 100 and 300.\footnote{\url{https://radimrehurek.com/gensim/models/word2vec.html} (Accessed 23 July 2025)} In contrast, a \gls{bert} model is a \gls{dnn} with several hidden layers (e.g., 12 layers), that learns the meaning of a word in a sentence depending on the surrounding words (context) and produces larger feature vectors, for example, of length 768.\footnote{\url{https://github.com/google-research/bert} (Accessed 23 July 2025)}

Further, the timestamp of each event is encoded. For that, first, one or more numerical time features are created from the timestamp that are suitable for transfer. An example of such a time feature is \textit{duration since start}, indicating the temporal difference from the first to the current event of a trace. After feature creation, the values of each feature are divided by a divisor, which is defined for the source domain and has an equivalent meaning in the target domain. For example, such a divisor could be the $x$-th quantile, the mean, or the maximum value of a time feature. With a divisor of this kind, the values of a time feature in the source domain are aligned to the values of the corresponding time feature in the target domain.  

Third, prefix encoding is applied to the traces of the event log $L_S$. 
A \emph{prefix} is a sub-sequence of a trace or the complete trace itself. Prefixes are created to build and apply a base model, which can predict targets (e.g., a process-related outcome) for each time step in a running business process. 
For the prefix encoding, we apply the index-based sequence encoding approach~\citep{leontjeva2015complex}. In this approach, all information in the prefix (including the order) is used and features for each attribute of each executed event (index) are generated~\citep {Teinemaa.2019}. We encode prefixes in this way because we employ an \gls{lstm}-based model as base model, which can effectively learn temporal dependencies from the event data of a prefix, from the first to the current event.   

Fourth, the prefixes of the event log $L_S$ are transformed into the event data $\mathsf{X}_S \in \mathrm{R}^{s \times T \times v}$ and the label data $\textbf{y}_S \in \mathrm{R}^{s}$ to build a \gls{dnn}-based base model. Here, $s$ is the number of prefixes, $T$ refers to the number of time steps of the longest prefix, and $v$ denotes the number of features.
For the event data $\mathsf{X}_S$, the remaining space for prefixes comprising fewer time steps than the prefix with the maximum number of time steps is padded with zeros. For each time step of each prefix, $\mathsf{X}_S$ stores information on the activity and the timestamp attribute in the form of extracted features. The concatenation of the activity features and the time features represents the final feature vector with its size corresponding to the number of features $v$ of the event data $\mathsf{X}_S$. 
For the label data, labels are set on the trace level to predict target values as early as possible. This is common in outcome-oriented \gls{ppm} \citep{Teinemaa.2019}, and applicable in our work, as we focus on binary, outcome-based process predictions. 
For example, if the prediction target of interest is \textit{in-time}, the model is applied at each time step of a running process instance to predict whether this instance will be finished in time or not. 
Finally, the event data $\mathsf{X}_S$ and the label data $\mathbf{y}_S$ are split into a training set $(\mathsf{X}_S^{tr}, \mathbf{y}_{S}^{tr})$ and a validation set $(\mathsf{X}_S^{val}, \mathbf{y}_{S}^{val})$ for model training, and a test set $(\mathsf{X}_S^{te}, \mathbf{y}_S^{te})$ for model evaluation.

Fifth, given the training event data of the source context $\mathsf{X}_S^{tr}=(\mathbf{X}_1, \dots, \mathbf{X}_s)$, with $\mathbf{X}_i \in \mathrm{R}^{T \times v}$, the base model $f_{\text{DNN}}: \mathbf{X}_i \rightarrow \hat{y}_i \in (0,1)$ is created and trained, mapping event data instances $\mathbf{X}_i = (\mathbf{x}_1, \dots, \mathbf{x}_T)$, with $\textbf{x}_t \in \mathrm{R}^{v}$, onto estimations $\hat{y}_i$.    
%
The base model is a two-layer \gls{lstm} model. For each layer $l$ = \{1,2\} and time step $t=1, \dots, T$, the \gls{lstm}~\citep{hochreiter.1997}~transition is

\begin{align}
\label{eq:lstm}
\mathbf{f}_{g_{t}}^{(l)} & =sigmoid\left(\mathbf{U}_f^{(l)} \mathbf{h}_{t-1}^{(l)}+\mathbf{W}_f^{(l)} \mathbf{x}_{t}^{(l)}+\mathbf{b}_f^{(l)}\right) & \text { (forget gate), } \\
\mathbf{i}_{g_t}^{(l)} & =sigmoid\left(\mathbf{U}_i^{(l)} \mathbf{h}_{t-1}^{(l)}+\mathbf{W}_i^{(l)} \mathbf{x}_{t}^{(l)}+\mathbf{b}_i^{(l)}\right) & \text { (input gate), } \\
\tilde{\mathbf{c}}_{t}^{(l)} & =tanh \left(\mathbf{U}_g^{(l)} \mathbf{h}_{(t-1)}^{(l)}+\mathbf{W}_g^{(l)} \mathbf{x}_{t}^{(l)}+\mathbf{b}_g^{(l)}\right) & \text { (candidate memory), } \\
\mathbf{c}_{t}^{(l)} & =\mathbf{f}_{g_{t}}^{(l)} \circ \mathbf{c}_{t-1}^{(l)}+\mathbf{i}_{g_{t}}^{(l)} \circ \tilde{\mathbf{c}}_{t}^{(l)} & \text { (current memory), } \\
\mathbf{o}_{g_{t}}^{(l)} & =sigmoid\left(\mathbf{U}_o^{(l)} \mathbf{h}_{t-1}^{(l)}+\mathbf{W}_o^{(l)} \mathbf{x}_{t}^{(l)}+\mathbf{b}_o^{(l)}\right) & \text { (output gate), } \\
\mathbf{h}_{t}^{(l)} & =\mathbf{o}_{g_{t}}^{(l)} \circ tanh \left(\mathbf{c}_{t}^{(l)}\right) & \text { (current hidden state).}
\label{eq:lstm2}
\end{align}

In Eqs.~(\ref{eq:lstm}--\ref{eq:lstm2}), $\mathbf{U}_{\bigoplus}$, $\mathbf{W}_{\bigoplus}$, and $\mathbf{b}_{\bigoplus}$ for $\bigoplus \in \{f,i,g,o\}$ are trainable parameters, $\circ$ is the element-wise product, $\mathbf{h}_t$ and $\mathbf{c}_t$ are the hidden state and cell memory of the \gls{lstm} layer $l$. Stacking the two layers yields an \gls{lstm} encoder

\begin{equation}
    \Big(\mathbf{h}_T^{(2)},\mathbf{c}_T^{(2)} \Big) = f_{\text{LSTM}}\Big(\mathbf{X}_i \Big),
\end{equation}

where $\mathbf{h}_T^{(2)}$ and $\mathbf{c}_T^{(2)}$ is the last hidden state of the second layer. 
We use two \gls{lstm} layers in the base model to learn a multi-level hierarchy of features. The first layer learns features, capturing lower-level, temporal patterns, whereas the second layer, receiving a sequence of these features, can learn features with higher-level abstraction. The learning of such a multi-level hierarchy of features from event data of the source context is necessary in our setting to facilitate the knowledge transfer to the target context.

Then, the hidden state $\mathbf{h}_T^{(2)}$ is taken, projected through a fully-connected layer, and a sigmoid activation is applied to the projection's output to calculate the estimations, as formalized in~Eq.~(\ref{eq:out}).

\begin{equation}
\label{eq:out}
\hat{y}= sigmoid \Big(W_{\text {out }} \mathbf{h}_T^{(2)}+\mathbf{b}_{\text {out }}\Big).
\end{equation}

Before optimizing the base model's internal parameters, they are initialized to improve learning from the event data of the source context and to facilitate knowledge transfer to the target context. All input weight matrices $\mathbf{W}_{\bigoplus}$ for $\bigoplus \in \{f,i,g,o\}$ are Xavier-uniform initialized to center the values around 0, going into sigmoid/tanh activations. This enables the \gls{lstm} to start in a flexible area and encourages the optimizer to obtain useful gradients immediately. 
All recurrent weight matrices $\mathbf{U}_{\bigoplus}$ for $\bigoplus \in \{f,i,g,o\}$ are initialized with a random orthogonal matrix to keep the singular values of the recurrent map near 1 and to reduce the problem of exploiting/vanishing gradients~\citep{arjovsky2016unitary}. In the recurrent part of the \gls{lstm}, the recurrent map is the operation that takes the hidden state at $t-1$ and transforms it to the next hidden state at time $t$, using the recurrent weight matrices. Singular values of the recurrent weight matrices measure how much the recurrent map stretches or squashes vectors in different directions. For example, if a singular value is $>1$, that direction grows exponentially over time (i.e., gradients explode).  
The forget-gate bias $\mathbf{b}_f$ is set to +1 (other LSTM biases $\{\mathbf{b}_i, \mathbf{b}_g, \mathbf{b}_o\}$ are set to 0) to encourage the network to preserve memory at the start of training~\citep{jozefowicz2015empirical}. 
For the final layer, the weights $\mathbf{W}_{out}$ are He-uniform initialized with 0 bias to allow for more stability in training, given the fact that our model has one output neuron to model the two possible process-related outcomes (e.g., in-time or not), which our technique assumes.

The internal parameters of the base model are optimized by solving the following optimization problem:
\begin{align}
 \beta^*_{S} &=\underset{\beta_{S}}{\arg \min}\sum^{s}_{i=1}\pazocal{L}\Big(f_{{\text{DNN}}}\Big(\mathbf{X}_{S, i}^{tr}; \beta_S\Big), y_{S,i}^{tr}\Big),
\end{align}

where $\pazocal{L}$ is a \gls{bce} function, $\beta^{*}_{S}$ are the adjusted internal parameters of the model after training on training event data of the source context $S$, $s$ is the number of prefixes, $\mathbf{X}_{S, i}^{tr}$ is the i-th prefix of the training event data $\mathsf{X}^{tr}_S$ of the source context~$S$, and $y_{S, i}^{tr}$ is the i-th label value of training label data $\mathbf{y}_{S}^{tr}$ of the source context~$S$.  
During the internal parameter optimization, early stopping is performed if no improvement in the \gls{bce} on the validation set $(\mathsf{X}_S^{val}, \mathbf{y}_S^{val})$ takes place over 10 subsequent epochs.

Sixth, the built base model $f_{\text{DNN}}(\cdot)$ is evaluated to determine whether it should be transferred from the source context $S$ to the target context $T$. 
To do that, the base model is applied to the test event data $(\mathsf{X}_S^{te}, \mathbf{y}_S^{te})$ and its prediction performance is measured via \gls{ml} metrics $m \in M$; that is, $m(u(f_{\text{DNN}}(\mathsf{X}_S^{te};\beta^*_{S})), \mathbf{y}_S^{te})$, where $u$ is a function transforming model estimations to class predictions or probabilities. As \gls{ml} metrics, commonly used in outcome-oriented \gls{ppm} are calculated, such as the $\text{AUC}_{\text{ROC}}$ and the F1-score~\citep{Teinemaa.2019}. In addition, the prediction performance, measured via the \gls{ml} metrics, should be considered depending on the used validation strategy, as it determines how the event data is split for model evaluation. 

\subsection{Transfer resources from source to target}

The second phase, which aims to transfer relevant and usable resources from the source context $S$ to the target context $T$, consists of two steps. 

First, relevant resources are transferred from the source context to the target context. These resources can be grouped into \gls{tl}-specific resources and general resources. \gls{tl}-specific resources include the pre-trained embedding model for activity encoding, information for timestamp encoding (one or more selected time features for transfer and divisors for feature value scaling), and the pre-trained prediction model for application. General resources include further information on pre-processing (e.g., the type of prefix encoding) and evaluation (e.g., \gls{ml} metrics for measuring the base model's prediction performance).   

Second, if traces in the form of an event log $L_T$ are available of the target context $T$, the base model $f_{\text{DNN}}(\cdot)$ from the source context $S$ can be applied and evaluated in an intermediary step before it is used to create class predictions and probabilities based on (ongoing) process instances in the target context $T$. 
In this intermediary step, first, the event log $L_T$ is loaded, and activity and timestamp encoding are performed. For activity encoding, the transferred, pre-trained embedding model $f_{\text{EMB}}(\cdot)$ is applied, whereas for timestamp encoding, the transferred information on time feature selection and scaling is used. 
After activity and timestamp encoding, prefixes are encoded, the event data $\mathsf{X}_T$ and the label data $\mathbf{y}_T$ are created, and the data are split into a training set $(\mathsf{X}_T^{tr}, \mathbf{y}_{T}^{tr})$, a validation set $(\mathsf{X}_T^{val}, \mathbf{y}_T^{val})$, and a test set $(\mathsf{X}_T^{te}, \mathbf{y}_T^{te})$. 
To ensure that the pre-processing is performed in the same way as in the source context $S$, further pre-processing information is transferred from the source context $S$ to the target context $T$.
Moreover, the transferred model $f_{\text{DNN}}(\cdot)$ could be fine-tuned with event data of the target context $T$ (i.e., $(\mathsf{X}_T^{tr}, \mathbf{y}_{T}^{tr})$ and $(\mathsf{X}_T^{val}, \mathbf{y}_T^{val})$). However, by default, the base model is not fine-tuned with any event data of the target context. 
Finally, the transferred base model is applied to the test set $(\mathsf{X}_T^{te},\mathbf{y}_T^{te})$ to evaluate its prediction performance.  
For this purpose, evaluation information, such as used \gls{ml} metrics, is transferred from the source context $S$ to the target context $T$ to assess the prediction performance of the base model in the same way as in the source context. 

For the evaluation in this intermediary step, there are two further aspects to consider. 
First, as the test sets of the event logs from the target and source contexts are naturally different, the test \gls{ml} metrics cannot be directly compared with each other. To overcome that, a second prediction model should be trained \textquote{from scratch} using the training set $(\mathsf{X}_T^{tr}, \mathbf{y}_{T}^{tr})$ of the target context $T$. Based on the test set $(\mathsf{X}_T^{te}, \mathbf{y}_T^{te})$, this second prediction model is compared with the transferred base model in terms of prediction performance to evaluate the performance of the transfer. 
Second, if the event log $L_T$ only comprises a few traces, the entire event log should be used as a test set in the evaluation of the base model to get an initial indication of how well the base model performs in the target context.

\subsection{Online application of model to target}

The third phase, which aims to create class predictions and probabilities for a new, ongoing process instance in the target context $T$ based on the resources transferred from the source context $S$, consists of six steps. 

First, the (ongoing) process instance $\sigma_T$ is loaded, which represents a prefix of a trace that will be completed in the future. 
Then, activity encoding and timestamp encoding are performed on the (ongoing) process instance. For the activity encoding, the transferred, pre-trained embedding model $f_{\text{EMB}}(\cdot)$ is applied, and for timestamp encoding, the transferred information on time feature selection and scaling is used.
In the next step, the prefix is encoded before it is transformed into the event data instance $\mathbf{X}_T$. To also ensure that the pre-processing in the online application is performed in the same way as in the source context $S$, further pre-processing information is transferred from the source context $S$ to the target context $T$.   
%
Then, the transferred base model $f_{\text{DNN}}(\cdot)$ is applied to the event data instance $\mathbf{X}_T$ to produce the estimate $\hat{y}$, which is transformed into a class prediction and distribution. In contrast to the previous two phases, the prediction performance is not evaluated, as no label information is available in the online application. 
Instead, in this phase, the class prediction and distribution are used for operational support by process users or other stakeholders to, for example, proactively draw attention to undesired process outcomes, with the overall objective of improving the performance of the underlying business process. 

\section{Evaluation}
\label{sec:evaluation}

To evaluate our technique, we instantiated it in two real-life use cases. 
The first one represents an intra-organizational use case, whereas the second one is an inter-organizational use case. In doing so, we account for the differences between transfer scenarios and can thoroughly evaluate the performance of our technique for both intra- and inter-organizational scenarios. For example, in an intra-organizational use case where the application of an existing prediction model is extended to a new product or service within the same process context, the performed activities can be similar, while variations may occur in other dimensions, such as the sequence of activities or throughput time. In an inter-organizational use case, the performed activities might even vary in their descriptions. 

For each use case, we evaluated the technique using the corresponding instantiation in five regards. 
First, we evaluated the overall prediction performance of the technique by comparing it to that of traditional \gls{ppm} techniques. 
Second, we evaluated the effect of applying different pre-trained models for activity embedding in our technique on the prediction performance.
Third, we evaluated the embedding difference between the activities from the source and the target context. 
Fourth, we evaluated the effect of the proposed relative cross-domain approach for timestamp encoding on the prediction performance compared to other timestamp encoding techniques. 
Fifth, we evaluated the transfer relevance by comparing the prediction performance of the transferred model with that of models only trained with varying amounts of training event data of the target context.
In addition, we conducted further evaluations to assess the prediction performance of our \gls{tl}-based \gls{ppm} technique i) when the complete model is used versus sub-variants of it (see~Appendix~\ref{app:SourceLogScaling}), ii) when different prefix encoding approaches are applied (see Appendix~\ref{app:PrefixEncodingTechniques}), and iii) when activity and timestamp information about prefixes are encoded separately versus jointly~(see~Appendix \ref{app:PrefixIsoJointPrefixEncoding}).

\subsection{Use case descriptions}

\subsubsection{Use case I: Intra-organizational transfer}
\label{sec:eval:usecaseI}

Intra-organizational \gls{tl} allows an organization to transfer knowledge across its business processes. To evaluate our \gls{tl}-based \gls{ppm} technique in the intra-organizational use case, we applied it to process data handling different products. In particular, we used the event data of two sub-logs from one organization, specifically sub-logs of the \textit{bpic2014} event log\footnote{\url{https://data.4tu.nl/articles/_/12692378/1} (Accessed 23 July 2025)}. The data were provided by an \gls{ict} company from the Netherlands, called Rabobank Group ICT. It concerns their \gls{itsm} processes, specifically their incident management process. The data were derived from the company's \gls{itil} service management tool, called the HP Service Manager, and refer to incidents that are recorded when a reported problem cannot be resolved over the phone. Each incident is related to a service component, that is, one particular product in the bill of materials.
The first sub-log \textit{WBS72\&223}, the source log, only includes traces related to the service components \textit{WBS000072} and \textit{WBS000223}. The second sub-log \textit{WBS263}, our target log, only includes traces related to the service component \textit{WBS000263}. In the pre-processing, traces longer than 50 events were removed.

Although both event logs are from the same organization, each service component can be handled differently. For example, the distribution of organizational teams assigned to the activities in both logs varies between service components. This different handling is reflected in different event log characteristics in terms of size, number of variants, and activities, which is why the processes are different. 
The characteristics of both event logs after pre-processing are summarized in Table~\ref{tab:logs_intra}.

\begin{table}[ht]
\centering
\caption{Characteristics of the event logs used for the intra-organizational use case.}
\label{tab:logs_intra}
\small
\begin{tabular}{lrr}
\toprule
                          &\multicolumn{2}{c}{\textbf{Characteristics}}\\
\textbf{Number of} & \textbf{WBS72\&223} ($L_S$)   & \textbf{WBS263} ($L_T$)  \\ \midrule
Instances      &   2,959           &  2,204         \\
Instance Variants      & 1,599      &  1,476           \\
Events    &          28,151 &   26,692           \\
Activities     &         26    &  29             \\ 
Events per Instance*  &{[}3; 50; 9.51{]} &{[}3; 50; 12.11{]}\\
Throughput time in days*  &{[}0.0002; 167.86; 5.83{]} &{[}0.0004; 224.94; 4.68{]}\\
\bottomrule
\multicolumn{3}{l}{* {[}min; max; mean{]}}
\end{tabular}

\end{table}

An overview of the activities and their distribution can be found in Table~\ref{tab:actslogs_intra}.

\begin{table}[ht]
\centering
\caption{Activities and their frequency of the event logs used for the intra-organizational use case.}
\label{tab:actslogs_intra}
\small
\begin{tabular}{@{}lrllr@{}}
\toprule
\multicolumn{2}{c}{\textbf{WBS72\&223 ($L_S$)}} &&\multicolumn{2}{c}{\textbf{WBS263 ($L_T$)}} \\ 
 \textbf{Activity}                                   & \textbf{Frequency} &&\textbf{Activity}& \textbf{Frequency} \\ \cline{1-2} \cline{4-5}
Assignment & 5.336 && Assignment & 6.376 \\
Reassignment & 4.504 && Reassignment & 5.131 \\
Update & 4.259 && Update & 4.146 \\
Closed & 3.343 && Operator Update & 2.532 \\
Open & 2.959 && Closed & 2.349 \\
Operator Update & 2.758 && Open & 2.204 \\
Status Change & 1.953 && Caused By CI & 2.044 \\
Caused By CI & 1.718 && Status Change & 777 \\
Reopen & 320 && Description Update & 309 \\
Mail to Customer & 185 && Impact Change & 154 \\

\bottomrule
\multicolumn{5}{l}{\textit{Note.} Only the ten most frequent activities are listed.}
\end{tabular}

\end{table}

For our evaluation, we predict whether running process instances will be finished in time or not. The label was constructed by taking the 70\textsuperscript{th} quantile of all traces' duration as the criterion for \textit{in-time} for each sub-log independently. The value of the criterion for the source log \emph{WBS72\&223} was 4.94 days, and for the target log \emph{WBS263}, 4.15 days. Traces with a longer duration were labeled as \textit{not in-time}.

\subsubsection{Use case II: Inter-organizational transfer}
\label{sec:eval:usecaseII}

In inter-organizational \gls{tl}, process knowledge is transferred across organizational boundaries. For the evaluation of our technique in the inter-organizational use case, we use two event logs from different organizations. The first one is the \textit{bpic2014} event log\footnote{\url{https://data.4tu.nl/articles/_/12692378/1} (Accessed 23 July 2025)}. The second event log called \textit{helpdesk}\footnote{\url{https://data.4tu.nl/articles/dataset/Dataset_belonging_to_the_help_desk_log_of_an_Italian_Company/12675977} (Accessed 23 July 2025)} covers the same type of process, an \gls{it} ticket management process.
Both event logs were derived from a ticketing system and, therefore, are comparable in their structure and notation. In our scenario for \gls{tl}, the \textit{bpic2014} serves as the source log, and the \textit{helpdesk} event log as the target log. As the event logs from the different organizations exhibit different characteristics in terms of size, number of variants, and activities, the processes are different.

The characteristics of both event logs after the pre-processing are summarized in Table~\ref{tab:logs_inter}. As with the intra-organizational use case, traces longer than 50 events were removed.

\begin{table}[ht]
\centering
\caption{Characteristics of the event logs used for the inter-organizational use case.}
\label{tab:logs_inter}
\small
\begin{tabular}{lrr}
\toprule
                           &\multicolumn{2}{c}{\textbf{Characteristics}}\\
\textbf{Number of} & {\textbf{bpic2014 ($L_S$)}} & {\textbf{helpdesk ($L_T$)}}   \\ \midrule
Instances &  46,174     &  4,480                    \\
Instance Variants & 22,190         &  226               \\
Events &  435,643       &          21,348           \\
Activities &     38       &   14                 \\ 
Events per Instance* &{[}1; 50; 9.43{]} &{[}2; 15; 4.66{]}\\
Throughput time in days*  &{[}0.0; 392.06; 4.87{]} &{[}30.64; 59.99; 40.86{]}\\
\bottomrule
\multicolumn{3}{l}{* {[}min; max; mean{]}}
\end{tabular}

\end{table}

An overview of the activities and their distribution can be found in Table~\ref{tab:actslogs_inter}.

\begin{table}[ht]

\centering
\caption{Activities and their frequency of the event logs used for the inter-organizational use case.}
\label{tab:actslogs_inter}
\small
\begin{tabular}{@{}lrllr@{}}
\toprule
\multicolumn{2}{c}{\textbf{bpic2014 ($L_S$)}} &&\multicolumn{2}{c}{\textbf{helpdesk ($L_T$)}} \\ 
 \textbf{Activity}                                   & \textbf{Frequency} &&\textbf{Activity}& \textbf{Frequency} \\ \cline{1-2} \cline{4-5}
Assignment & 80.123 && Take in charge ticket & 5.060 \\
Operator Update & 50.563 && Resolve ticket & 4.983 \\
Closed & 49.270 && Assign seriousness & 4.938 \\
Status Change & 48.423 && Closed & 4.574 \\
Open & 46.161 && Wait & 1.463 \\
Reassignment & 45.553 && Require upgrade & 119 \\
Caused By CI & 34.090 && Insert ticket & 118 \\
Update & 32.348 && Create SW anomaly & 67 \\
Quality Indicator Fixed & 7.651 && Resolve SW anomaly & 13 \\
Communication with customer & 5.675 && Schedule intervention & 5 \\
Description Update & 4.285 && VERIFIED & 3 \\
Pending vendor & 4.188 && RESOLVED & 2 \\
External Vendor Assignment & 4.145 && INVALID & 2 \\
Mail to Customer & 3.763 && DUPLICATE & 1 \\
Update from customer & 3.235 &&  & \\

\bottomrule
\multicolumn{5}{l}{\textit{Note.} Only the 15 most frequent activities are listed.}
\end{tabular}

\end{table}

Just like the intra-organizational use case, we predict whether running process instances will be finished in time or not. The label was constructed by taking the 70\textsuperscript{th} quantile of all traces' duration as the criterion for \textit{in-time} for each log independently. The value of the criterion for the source log \emph{bpic2014} was 2.83 days, and for the target log \emph{helpdesk}, 44.95 days. Traces with a longer duration were labeled as \textit{not in-time}.

\subsection{Setup}

For our first three evaluations on the prediction performance of our technique, we sorted the process instances of the event logs of the source and target contexts in temporal order according to the timestamps of their first event. We then split each event log into a 64\% training, a 16\% validation, and a 20\% test set, trained the models, and computed the $\text{AUC}_{\text{ROC}}$ and the \gls{mcc}, as well as the weighted precision, recall, and F1-score based on the prediction results on test sets. Among these \gls{ml} metrics, we choose the $\text{AUC}_{\text{ROC}}$ as the primary metric because it is independent of the threshold, which, therefore, allows for a more holistic evaluation of the models' prediction performance. We repeated each experiment five times with different seeds and calculated the average values and the standard deviation across the five runs. 
In addition, for each approach tested in the first three evaluations, we used two additional baselines and compared their prediction performance with the model of the respective approach trained on the event data of the source context and transferred to the target context without fine-tuning. 
The first baseline is a model trained and tested on the event data of the source context, which serves as a model with a reference prediction performance in the source context. 
The second baseline is a model trained and tested on the event data of the target context, which serves as a model with a reference prediction performance in the target context. 

For the \textit{first} evaluation, in which we compared the prediction performance of our \gls{tl}-based technique to that of the traditional \gls{ppm} approaches, we selected five baseline approaches. 
The first baseline employs the same \gls{lstm} model as in our proposed technique.
To encode the activities for this and the other models, one-hot encoding is applied. It is a common technique for encoding categorical attributes of events in \gls{ppm}~\citep[e.g.,][]{rama2021deep}.
However, since we do not know the target context's activities when training the base model in the source context, only the source context's activities are used to fit the one-hot encoder. The encoder is then applied to transform the target context's activities. Consequently, if an activity exists in both contexts, it can be encoded in the target context's event data; otherwise, this is not possible.
In addition, for the \emph{duration since start} time feature, a min-max scaler is only fitted on the training event data from the source context and applied to the test event data from the source and target contexts to avoid data leakage.   
The second baseline employs a logistic regression model with the same encoding strategy as the first approach.
The last three baselines use either an \gls{xgb}, a random forest, or a decision tree model. The activities are one-hot encoded again, while the \emph{duration since start} time feature is not scaled or normalized. This is because the underlying algorithms for building these models are tree-based and split data based on feature thresholds rather than the absolute magnitude or distribution of features. The models employed in all baselines are commonly used in outcome-oriented \gls{ppm}~\citep{Teinemaa.2019,maggi2014predictive,wang2019outcome}.

For the \textit{second} evaluation, in which we investigated the effect of using different pre-trained models for activity encoding on prediction performance, three types of embedding models were employed. 
The first group includes three static word embedding models. The first two models, \textit{glove-wiki-gigaword-100} and \textit{glove-wiki-gigaword-300}, are \gls{glove} models\footnote{\url{https://radimrehurek.com/gensim/models/word2vec.html\#other-embeddings} (Accessed 23 July 2025)} that output feature vectors of length 100 and 300 for the activities, respectively.
The third model, \textit{word2vec-google-news-300}, is a pre-trained word2vec model\footnote{\url{https://radimrehurek.com/gensim/models/word2vec.html} (Accessed 23 July 2025)} that encodes activities as vectors of length 300. 
Although these encoding models have the limitation that all input words must be included in the vocabulary of the pre-trained model, we did not replace any domain-specific acronyms included in the activity names to ensure a fair comparison. For example, we did not replace the acronym \textquote{CI} with \textquote{configuration item} in the activity name \textquote{Caused By CI} of the bpic2014 event log. 
The second group includes two contextual word embedding models. The first model is \emph{bert-base-cased}\footnote{\url{https://huggingface.co/google-bert/bert-base-cased} (Accessed 23 July 2025)}, a pre-trained \gls{bert} model \citep{kenton2019bert}. 
This \gls{bert} model has the same architecture as the original \gls{bert} model, which is entirely based on the encoder stack of the transformer model \citep{vaswani2017attention}. The pre-training of this model was performed on a large corpus using the masked language modeling (prediction of randomly masked words in sentences based on the surrounding words) and the next sentence prediction objectives.  %
The second model is \emph{bert-base-uncased}\footnote{\url{https://huggingface.co/google-bert/bert-base-uncased} (Accessed 23 July 2025)}, another pre-trained \gls{bert} model, which differs from the \emph{bert-base-cased} model by lower-casing all input text before tokenization.  
Both models output contextualized word embeddings of length 748 for each token. As activities typically consist of several tokens, a mean pooling operation is performed on top of the contextual word embeddings to produce one embedding vector per activity. 
The third group includes two contextual sentence embedding models.
The first model is \emph{all-MiniLM-L12-v2}\footnote{\url{https://huggingface.co/sentence-transformers/all-MiniLM-L12-v2} (Accessed 23 July 2025)}, a fine-tuned version of the pre-trained Mini-LM model \citep{wang2020minilm}~\emph{MiniLM-L12-H384-uncased}\footnote{\url{https://huggingface.co/sentence-transformers/all-MiniLM-L12-v2} (Accessed 23 July 2025)}. Fine-tuning is performed using a contrastive learning objective (predicting for a sentence the correct matching sentence from a group of randomly selected sentences), which allows for a better capture of the semantic meaning of entire sentences.    
The second model is \emph{all-mpnet-base-v2}\footnote{\url{https://huggingface.co/sentence-transformers/all-mpnet-base-v2} (Accessed 23 July 2025)}, a version of the pre-trained \emph{mpnet} model \citep{song2020mpnet} for which fine-tuning was also performed using a contrastive learning objective.
The \emph{all-MiniLM-L12-v2} and \emph{all-mpnet-base-v2} models output contextualized word embeddings of lengths 384 and 768, respectively, and, as with the contextual word embedding models, a mean pool operation was performed to get one embedding vector per activity.

For the \textit{third} evaluation, which assesses the quality of the activity embeddings, we examine the similarity of the activity embeddings from the source and target contexts for the intra- and inter-organizational use cases. For this evaluation, we used the embedding models that demonstrated the highest test $\text{AUC}_{\text{ROC}}$ scores in the target context, calculated the Euclidean distance between the embedding vectors, and visualized the distances through a heatmap.

For the \textit{fourth} evaluation, in which we assessed the effect of the proposed relative cross-domain approach for timestamp encoding on the prediction performance, we employed two different baseline approaches.  
In the first approach, the time $h$ (in the form of the number of hours), the day of the week $d$, and the month $m$ are extracted from the timestamp and coded as separate features. These variables have a cyclical structure: Sunday is followed by Monday, December is followed by January, and the 23rd hour is followed by the 0-th hour. To take this periodic nature into account, the coding is carried out using a sinus function that can depict cyclical relationships.
For this purpose, the months are numbered from 0 (January) to 11 (December), the days of the week from 0 (Monday) to 6 (Sunday), and the hours from 0 to 23, and transformed as follows:
\begin{align}
d_{emb} &= sin\left(2\cdot\pi \cdot \frac{d}{7}\right), \\
m_{emb} &= sin\left(2\cdot\pi \cdot \frac{m}{12}\right), \\
h_{emb} &= sin\left(2\cdot\pi \cdot \frac{h}{24}\right).
\end{align}

This form of coding means, for example, that the values for Sunday and Monday are close together. The advantage of this periodic representation is that cyclical effects, such as seasonal fluctuations or differences between times of day, can be better captured by the model. In the present evaluation, all three features are tested both individually and in combination.
The second approach extracts the \emph{duration since start} time feature from the timestamp, trains an autoencoder model on this feature's values in the training event data of the source context, and applies the model to all values of this feature in the event data of the source and the target contexts. The idea behind the autoencoder model is to reconstruct the input values using an encoder and a decoder. The encoder is then used to transform the feature values into a compact form (i.e., the latent space).

For the \textit{fifth} evaluation, we assessed the transfer relevance by comparing the prediction performance of the transferred model with that of models only trained with varying amounts of training event data of the target context. For this evaluation, we used the embedding models that led to the highest test $\text{AUC}_{\text{ROC}}$ scores in the target context. 

The \gls{lstm} models trained in all experiments have hidden layers with an internal element size of 128. The \gls{adam} optimizer \citep{kingma2015adam} was employed for optimizing the models' internal parameters, with a learning rate of $\text{1e}^{-3}$ and a maximum number of epochs set to 100. All experiments were conducted on a workstation equipped with 12 CPU cores, 128 GB of RAM, and a single NVIDIA RTX 6000 graphics card. All used and produced materials of the experiments can be found in the online repository of this work.\footnote{\url{https://github.com/fau-is/tl4ppm}}

\section{Results}
\label{sec:Results}

\subsection{Use case I: Intra-organizational transfer}

For the intra-organizational use case, we evaluated the prediction performance of our technique by comparing it with traditional \gls{ppm} approaches. 
Furthermore, we investigated the feature-based transfer of our technique in terms of different activity embeddings, including the quality of embeddings, and different timestamp encodings. Lastly, the transfer relevance was examined.

\subsubsection{Prediction performance for transfer learning-based compared to traditional predictive process monitoring}
\label{sec:results:intra_trad}

Table~\ref{tab:Results_intra} shows the prediction performance of the proposed \gls{tl}-based \gls{ppm} technique and the two baseline approaches, compared to five traditional \gls{ppm} approaches, for the intra-organizational use case. We chose the best activity embedding (see Section~\ref{sec:results:intra_actemb}) and timestamp encoding (see Section~\ref{sec:results:intra_timenc}) for our proposed technique.
It becomes apparent that our proposed technique achieves the highest scores in terms of precision, F1-score, \gls{mcc}, and $\text{AUC}_{\text{ROC}}$ compared to the traditional \gls{ppm} approaches employing \gls{xgb}, random forest, decision tree, and \gls{lstm} models. 
The approach using the logistic regression model demonstrates a slightly superior recall and the second-highest F1-score among the evaluated approaches. The prediction performance closest to our proposed \gls{tl}-based \gls{ppm} technique in terms of \gls{mcc} and $\text{AUC}_{\text{ROC}}$ is the traditional \gls{ppm} technique employing the \gls{lstm} model.
Comparing the results of the transferred models with the baseline approach \emph{Train and test on $L_T$}, the transferred models of our technique, as well as the techniques using logistic regression, \gls{xgb}, and \gls{lstm} models, surpass the baseline approach \emph{Train and test on $L_T$} in most metrics. 

\begin{table}[ht!]
\centering
\caption{Prediction performance of our proposed \gls{tl}-based \gls{ppm} technique and traditional \gls{ppm} techniques for the intra-organizational use case (average and standard deviation over five runs).}
\label{tab:Results_intra}
\resizebox{\textwidth}{!}{ 
\begin{tabular}{@{}lrrrrr@{}}
\toprule
\textbf{}                                       \textbf{PPM technique}      & \textbf{Precision} & \textbf{Recall} & \textbf{F1-score} & \textbf{MCC} &\textbf{$\text{AUC}_{\text{ROC}}$} 
\\ \midrule

\multicolumn{6}{l}{\textsc{TL-based PPM Technique}}\\
\textbf{Our proposed technique}                       &                   &                   &                   &              &    \\
Train and test on $L_S$    & 0.676 ($\pm$.005)   & 0.702 ($\pm$.022)     &  0.682 ($\pm$.006)  &  0.151 ($\pm$.012) &  0.641 ($\pm$.012)\\
Train and test on $L_T$    & 0.698 ($\pm$.020)   & 0.707 ($\pm$.020)     &  0.697 ($\pm$.019)  &  0.216 ($\pm$.059) &  0.688 ($\pm$.021)\\
Train on $L_S$, test on $L_T$ & \textbf{0.707} ($\pm$.010) & 0.704 ($\pm$.024) & \textbf{0.702} ($\pm$.013) & \textbf{0.241} ($\pm$.027) & \textbf{0.703} ($\pm$.008) \\
\midrule 

\multicolumn{6}{l}{\textsc{Traditional PPM Techniques}}\\
\textbf{LSTM} & & & & & \\
Train and test on $L_S$    & 0.670 ($\pm$.005)   & 0.666 ($\pm$.013)     &  0.668 ($\pm$.008)  &  0.143 ($\pm$.014)&  0.640 ($\pm$.007)\\
Train and test on $L_T$           & 0.644 ($\pm$.010) &  0.704 ($\pm$.019)     &  0.655 ($\pm$.013)        & 0.070 ($\pm$.030) & 0.552 ($\pm$.022)\\
Train on $L_S$, test on $L_T$    & 0.704 ($\pm$.014)&  0.654 ($\pm$.016) &   0.670 ($\pm$.013) &   0.228 ($\pm$.030)       & 0.683 ($\pm$.022) \\

\textbf{Logistic Regression}                         &                   &                   &                   &              &    \\
Train and test on $L_S$    & 0.681 ($\pm$.000)   & 0.734 ($\pm$.000)     & 0.677 ($\pm$.000)  & 0.136 ($\pm$.000) & 0.651 ($\pm$.000)\\
Train and test on $L_T$    & 0.648 ($\pm$.000)   & 0.707 ($\pm$.000)     & 0.661 ($\pm$.000)  & 0.081 ($\pm$.000) & 0.630 ($\pm$.000)\\
Train on $L_S$, test on $L_T$ & 0.684 ($\pm$.000) & \textbf{0.724} ($\pm$.000)     & 0.690 ($\pm$.000)   & 0.170 ($\pm$.000)  & 0.627 ($\pm$.000)\\

\textbf{XGBoost}                      &                   &                   &                   &              &    \\
Train and test on $L_S$    & 0.641 ($\pm$.000)   & 0.691 ($\pm$.000)     & 0.657 ($\pm$.000)  & 0.062 ($\pm$.000) & 0.622 ($\pm$.000)\\
Train and test on $L_T$    & 0.651 ($\pm$.000)   & 0.702 ($\pm$.000)     & 0.664 ($\pm$.000)  & 0.091 ($\pm$.000) & 0.596 ($\pm$.000)\\
Train on $L_S$, test on $L_T$ & 0.663 ($\pm$.000) & 0.701 ($\pm$.000)   & 0.674 ($\pm$.000)  & 0.124 ($\pm$.000) & 0.599 ($\pm$.000)\\

\textbf{Random Forest}                      &                   &                   &                   &              &    \\
Train and test on $L_S$    & 0.645 ($\pm$.003)   & 0.672 ($\pm$.003)     & 0.656 ($\pm$.003)  & 0.078 ($\pm$.007) & 0.599 ($\pm$.004)\\
Train and test on $L_T$    & 0.660 ($\pm$.006)    & 0.696 ($\pm$.005)     & 0.672 ($\pm$.005)  & 0.118 ($\pm$.014) & 0.617 ($\pm$.004)\\
Train on $L_S$, test on $L_T$ & 0.653 ($\pm$.007) & 0.687 ($\pm$.006)     & 0.665 ($\pm$.006)  & 0.101 ($\pm$.018) & 0.600 ($\pm$.006)\\

\textbf{Decision Tree}                      &                   &                   &                   &              &    \\
Train and test on $L_S$    & 0.635 ($\pm$.004)   & 0.621 ($\pm$.006)     & 0.627 ($\pm$.005)  & 0.051 ($\pm$.010) & 0.548 ($\pm$.005)\\
Train and test on $L_T$    & 0.647 ($\pm$.011)   & 0.632 ($\pm$.013)     & 0.639 ($\pm$.012)  & 0.091 ($\pm$.028) & 0.546 ($\pm$.016)\\
Train on $L_S$, test on $L_T$ & 0.641 ($\pm$.002) & 0.623 ($\pm$.004)     & 0.631 ($\pm$.003)  & 0.075 ($\pm$.006) & 0.538 ($\pm$.003)\\

\bottomrule
\multicolumn{6}{p{18cm}}{\textit{Note.} Best results for training on $L_S$ and testing on $L_T$ are marked in bold.} 
\end{tabular}
}
\end{table}

\subsubsection{Prediction performance for different activity embedding models}
\label{sec:results:intra_actemb}

To find an effective activity embedding model for our proposed \gls{tl}-based technique for the intra-organizational use case, we explored seven activity embedding models. Table~\ref{tab:Results_intra_actemb} shows the results of these models, which can be grouped into three categories: i) static word, ii) contextual word, and iii) contextual sentence embedding models.
Embedding the activity attribute with the contextual sentence embedding model \emph{all-mpnet-base-v2} achieves the best results in all metrics, that is, precision, recall, F1-score, \gls{mcc}, and~$\text{AUC}_{\text{ROC}}$.

\begin{table}[ht!]
\centering
\caption{Prediction performance of our \gls{tl}-based \gls{ppm} technique with the proposed activity embedding created with different pre-trained embedding models for the intra-organizational use case (average and standard deviation over five runs).}
\label{tab:Results_intra_actemb}
\resizebox{\textwidth}{!}{ 
\begin{tabular}{@{}lrrrrr@{}}
\toprule
\textbf{}                                        \textbf{Embedding model}     & \textbf{Precision} & \textbf{Recall} & \textbf{F1-score} & \textbf{MCC} &\textbf{$\text{AUC}_{\text{ROC}}$}
\\ \midrule

\multicolumn{3}{l}{\textsc{Static Word Embedding Models}}\\

\textbf{glove-wiki-gigaword-100}                       &                  &                   &                   &              &    \\
Train and test on $L_S$    & 0.675 ($\pm$.008)   & 0.676 ($\pm$.051)     &  0.668 ($\pm$.026)  &  0.148 ($\pm$.018)&  0.634 ($\pm$.017)\\
Train and test on $L_T$    & 0.690 ($\pm$.012) &  0.699 ($\pm$.021)     &  0.689 ($\pm$.014)        & 0.194 ($\pm$.043) & 0.671 ($\pm$.024)\\
Train on $L_S$, test on $L_T$    & 0.697 ($\pm$.021)&  0.670 ($\pm$.045) &   0.674 ($\pm$.029) &   0.209 ($\pm$.047)       & 0.688 ($\pm$.014) \\

\textbf{glove-wiki-gigaword-300}                         &                   &                   &                   &              &    \\
Train and test on $L_S$    & 0.669 ($\pm$.008) & 0.658 ($\pm$.017) & 0.663 ($\pm$.012) & 0.139 ($\pm$.021) & 0.613 ($\pm$.023) \\
Train and test on $L_T$    & 0.693 ($\pm$.013) & 0.677 ($\pm$.016) & 0.684 ($\pm$.014) & 0.208 ($\pm$.033) & 0.660 ($\pm$.037) \\
Train on $L_S$, test on $L_T$ & 0.685 ($\pm$.023) & 0.642 ($\pm$.027) & 0.658 ($\pm$.025) & 0.183 ($\pm$.058) & 0.656 ($\pm$.036) \\

\textbf{word2vec-google-news-300}                      &                   &                   &                   &              &    \\
Train and test on $L_S$    & 0.661 ($\pm$.013)   & 0.674 ($\pm$.034)     &  0.660 ($\pm$.004)  &  0.113 ($\pm$.033) &  0.617 ($\pm$.015)\\
Train and test on $L_T$    & 0.706 ($\pm$.022)   & 0.713 ($\pm$.024)     &  0.708 ($\pm$.021)  &  0.240 ($\pm$.058) &  0.678 ($\pm$.037)\\
Train on $L_S$, test on $L_T$ & 0.693 ($\pm$.020) & 0.682 ($\pm$.038) &  0.678 ($\pm$.016) &  0.199 ($\pm$.050) &  0.672 ($\pm$.028)\\

\midrule

\multicolumn{3}{l}{\textsc{Contextual Word Embedding Models}}\\

\textbf{bert-base-cased}                      &                   &                   &                   &              &    \\
Train and test on $L_S$           & 0.669 ($\pm$.010) & 0.670 ($\pm$.021) & 0.667 ($\pm$.008) & 0.138 ($\pm$.025) & 0.632 ($\pm$.013)\\
Train and test on $L_T$           & 0.691 ($\pm$.009) & 0.699 ($\pm$.017) & 0.693 ($\pm$.011) & 0.201 ($\pm$.026) & 0.668 ($\pm$.018)\\
Train on $L_S$, test on $L_T$     & 0.685 ($\pm$.026) & 0.661 ($\pm$.041) & 0.666 ($\pm$.029) & 0.181 ($\pm$.061) & 0.667 ($\pm$.043)\\

\textbf{bert-base-uncased}                      &                   &                   &                   &              &    \\
Train and test on $L_S$           & 0.661 ($\pm$.008) & 0.680 ($\pm$.016) & 0.668 ($\pm$.006) & 0.116 ($\pm$.023) & 0.607 ($\pm$.014) \\
Train and test on $L_T$           & 0.688 ($\pm$.010) & 0.685 ($\pm$.034) & 0.671 ($\pm$.015) & 0.170 ($\pm$.040) & 0.644 ($\pm$.016) \\
Train on $L_S$, test on $L_T$     & 0.674 ($\pm$.028) & 0.659 ($\pm$.032) & 0.663 ($\pm$.026) & 0.156 ($\pm$.072) & 0.635 ($\pm$.032) \\

\midrule

\multicolumn{3}{l}{\textsc{Contextual Sentence Embedding Models}}\\

\textbf{all-MiniLM-L12-v2}                      &                   &                   &                   &              &    \\
Train and test on $L_S$           & 0.672 ($\pm$.011) & 0.674 ($\pm$.025) & 0.672 ($\pm$.017) & 0.148 ($\pm$.029) & 0.632 ($\pm$.011) \\
Train and test on $L_T$           & 0.701 ($\pm$.005) & 0.714 ($\pm$.022) & 0.703 ($\pm$.009) & 0.223 ($\pm$.011) & 0.684 ($\pm$.022) \\
Train on $L_S$, test on $L_T$   & 0.701 ($\pm$.016) & 0.657 ($\pm$.018) & 0.671 ($\pm$.013) & 0.221 ($\pm$.037) & 0.693 ($\pm$.015) \\

\textbf{all-mpnet-base-v2}                      &                   &                   &                   &              &    \\
Train and test on $L_S$    & 0.676 ($\pm$.005)   & 0.702 ($\pm$.022)     &  0.682 ($\pm$.006)  &  0.151 ($\pm$.012) &  0.641 ($\pm$.012)\\
Train and test on $L_T$    & 0.698 ($\pm$.020)   & 0.707 ($\pm$.020)     &  0.697 ($\pm$.019)  &  0.216 ($\pm$.059) &  0.688 ($\pm$.021)\\
Train on $L_S$, test on $L_T$ & \textbf{0.707} ($\pm$.010) & \textbf{0.704} ($\pm$.024) & \textbf{0.702} ($\pm$.013) & \textbf{0.241} ($\pm$.027) & \textbf{0.703} ($\pm$.008) \\

\bottomrule
\multicolumn{6}{p{18cm}}{\textit{Note.} Best results for training on $L_S$ and testing on $L_T$ are marked in bold.}

\end{tabular}
}
\end{table}

\subsubsection{Activity embedding quality}
\begin{figure}[ht]
\centering
\includegraphics[width=\textwidth]{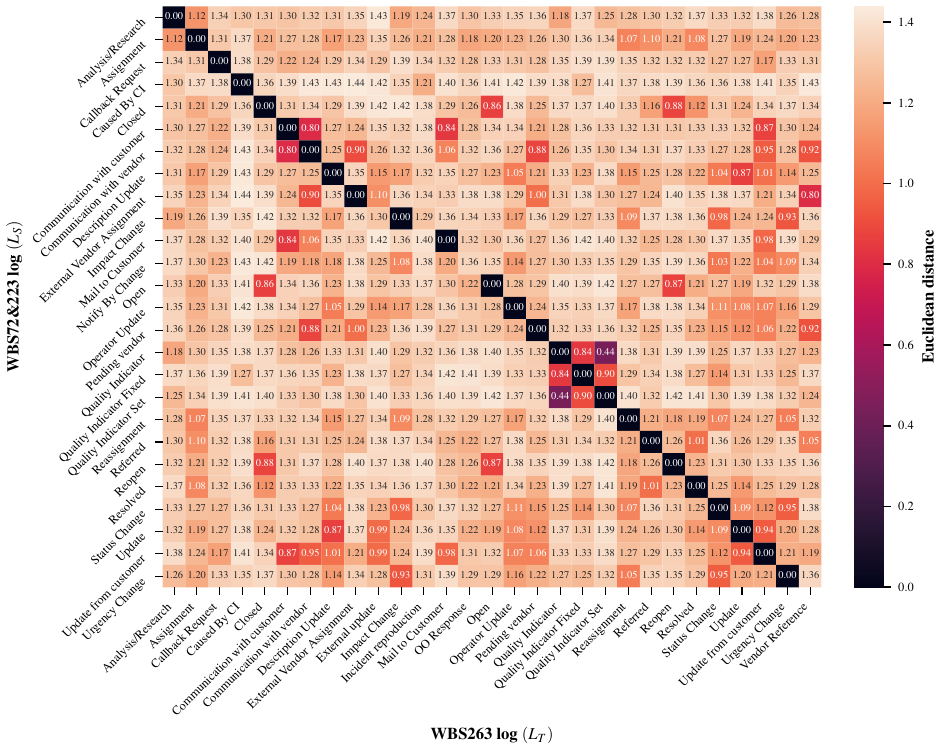}
\caption{Difference in embedding with model \emph{all-mpnet-base-v2} of activities in $L_S$ and $L_T$ for the intra-organizational use case.}
\label{fig:actem_intra} 
\end{figure}

To investigate the feature-based \gls{tl} part of our technique in more detail, we compared the difference in activity embeddings for the \emph{all-mpnet-base-v2} embedding model\footnote{The figures showing the Euclidean distance between activity embedding vectors created with the other embedding models of our setting can be found in the online repository of this work.} between $L_S$ and $L_T$ by calculating the Euclidean distance between the embedding vectors (see Figure~\ref{fig:actem_intra}).

The source and target logs have 25 activities in common, which can be seen by the black diagonal line formed by the 0.00 distances between the activity embeddings.
In addition, in both event logs, $L_S$ and $L_T$, there are two groups of activities (indicated by the deep-shaded squares), which contain activities that are recorded at a more granular level. The first group deals with communication, either \emph{Communication with customer} or \emph{Communication with vendor}.
The second group concerns the quality indicator, either referring to \emph{Quality Indicator} itself, or providing more details, as in \emph{Quality Indicator Fixed} or \emph{Quality Indicator Set}.
For both groups, the embeddings of the more granular activities are relatively similar, displaying a smaller distance in embedding vectors.

The target log $L_T$ has four additional activities that are not represented in the source log $L_S$. These are \emph{External update}, \emph{Incident reproduction}, \emph{OO Response}, and \emph{Vendor reference}.
\emph{External update} has a clear match with \emph{Update} and \emph{Update from customer} in $L_S$, and \emph{Vendor reference} with \emph{External vendor assignment} in $L_S$. For \emph{Incident reproduction} and \emph{OO Response}, the distances to the activity embeddings of $L_S$ are all similarly far apart. No clear match could be found based on the distances in embeddings.

\subsubsection{Prediction performance for different timestamp encoding approaches}
\label{sec:results:intra_timenc}

Besides investigating strategies to embed activities of our \gls{tl}-based \gls{ppm} technique, we examined five additional encoding strategies for the time-based feature \emph{duration since start} and compared them to our technique's relative cross-domain encoding. 
These five encoding strategies fall into two categories: i) time-based encoding and ii) autoencoder-based encoding.
The results in Table~\ref{tab:Results_intra_timenc} indicate a clear trend towards the relative cross-domain encoding as it achieves the highest precision, recall, F1-score, \gls{mcc}, and $\text{AUC}_{\text{ROC}}$ out of the six encoding strategies. The encoding strategy resulting in the second-best prediction performance values in all metrics is the autoencoder-based encoding.

\begin{table}[ht!]
\centering
\caption{Prediction performance of our \gls{tl}-based \gls{ppm} technique with the proposed relative cross-domain encoding of the \emph{duration since start} time feature and other encoding approaches for the intra-organizational use case (average and standard deviation over five runs).}
\label{tab:Results_intra_timenc}
\resizebox{\textwidth}{!}{ 
\begin{tabular}{@{}lrrrrr@{}}
\toprule
\textbf{}                                        \textbf{Encoding approach}     & \textbf{Precision} & \textbf{Recall} & \textbf{F1-score} &\textbf{MCC}   & \textbf{$\text{AUC}_{\text{ROC}}$}
\\ \midrule

\multicolumn{6}{l}{\textsc{Relative Cross-domain Encoding}}\\
Train and test on $L_S$    & 0.676 ($\pm$.005)   & 0.702 ($\pm$.022)     &  0.682 ($\pm$.006)  &  0.151 ($\pm$.012) &  0.641 ($\pm$.012)\\
Train and test on $L_T$    & 0.698 ($\pm$.020)   & 0.707 ($\pm$.020)     &  0.697 ($\pm$.019)  &  0.216 ($\pm$.059) &  0.688 ($\pm$.021)\\
Train on $L_S$, test on $L_T$ & \textbf{0.707} ($\pm$.010) & \textbf{0.704} ($\pm$.024) & \textbf{0.702} ($\pm$.013) & \textbf{0.241} ($\pm$.027) & \textbf{0.703} ($\pm$.008) \\
 
 \midrule

\multicolumn{6}{l}{\textsc{Time-based Encoding}}\\
\textbf{Hour}                       &                   &                   &                   &              &    \\
Train and test on $L_S$    &  0.654  ($\pm$.007)    &   0.672  ($\pm$.016)  &  0.660  ($\pm$.004) & 0.100  ($\pm$.020) & 0.612  ($\pm$.013) \\
Train and test on $L_T$           &  0.681  ($\pm$.011) & 0.685  ($\pm$.031)  & 0.681  ($\pm$.019) & 0.176  ($\pm$.029) & 0.642  ($\pm$.032) \\
Train on $L_S$, test on $L_T$    &    0.685  ($\pm$.012)       &  0.681  ($\pm$.020)         &   0.682  ($\pm$.015)        &   0.188  ($\pm$.031)  & 0.663 ($\pm$.014) \\ 
\textbf{Weekday}                       &                   &                   &                   &              &    \\
Train and test on $L_S$    &  0.662  ($\pm$.014)    &   0.658  ($\pm$.023)  &  0.658  ($\pm$.013) & 0.121  ($\pm$.036) & 0.612  ($\pm$.027) \\
Train and test on $L_T$           &  0.683  ($\pm$.007) & 0.678  ($\pm$.037)  & 0.675  ($\pm$.018) & 0.177  ($\pm$.017) & 0.654  ($\pm$.010) \\
Train on $L_S$, test on $L_T$    &    0.680  ($\pm$.018)       &  0.665  ($\pm$.031)         &   0.670  ($\pm$.023)        &   0.174  ($\pm$.047)  & 0.657 ($\pm$.033) \\ 
\textbf{Month}                       &                   &                   &                   &              &    \\
Train and test on $L_S$    &  0.650  ($\pm$.016)    &   0.661  ($\pm$.048)  &  0.651  ($\pm$.027) & 0.086  ($\pm$.038) & 0.592  ($\pm$.022) \\
Train and test on $L_T$           &  0.675  ($\pm$.021) & 0.693  ($\pm$.015)  & 0.679  ($\pm$.012) & 0.157  ($\pm$.056) & 0.630  ($\pm$.041) \\
Train on $L_S$, test on $L_T$    &    0.679  ($\pm$.019)       &  0.666  ($\pm$.043)         &   0.669  ($\pm$.031)        &   0.170  ($\pm$.047)  & 0.655 ($\pm$.033) \\ 
\textbf{Hour + Weekday + Month}                       &                   &                   &                   &              &    \\
Train and test on $L_S$    &  0.667  ($\pm$.009)    &   0.668  ($\pm$.023)  &  0.667  ($\pm$.016) & 0.134  ($\pm$.023) & 0.623  ($\pm$.011) \\
Train and test on $L_T$           &  0.655  ($\pm$.017) & 0.675  ($\pm$.021)  & 0.660  ($\pm$.008) & 0.106  ($\pm$.044) & 0.617  ($\pm$.025) \\
Train on $L_S$, test on $L_T$    &    0.695  ($\pm$.015)       &  0.676  ($\pm$.021)         &   0.682  ($\pm$.014)        &   0.210  ($\pm$.036)  & 0.676 ($\pm$.027) \\ 
\midrule

\multicolumn{6}{l}{\textsc{Autoencoder-based Encoding}}\\
Train and test on $L_S$    & 0.665 ($\pm$.013)   & 0.684 ($\pm$.023)     &  0.669 ($\pm$.007)  &  0.125 ($\pm$.036)&  0.630 ($\pm$.012)\\
Train and test on $L_T$    & 0.692 ($\pm$.014) &  0.700 ($\pm$.032)     &  0.694 ($\pm$.022)        & 0.204 ($\pm$.036) & 0.665 ($\pm$.020)\\
Train on $L_S$, test on $L_T$    & 0.703 ($\pm$.009)&  0.699 ($\pm$.019) &   0.698 ($\pm$.009) &   0.230 ($\pm$.025)    & 0.699 ($\pm$.013) \\

\bottomrule
\multicolumn{6}{p{18cm}}{\textit{Note.} Best results for training on $L_S$ and testing on $L_T$ are marked in bold.}

\end{tabular}
}
\end{table}

\subsubsection{Transfer relevance}
\begin{figure}[ht] 
\centering
	\includegraphics[width=0.8\textwidth]{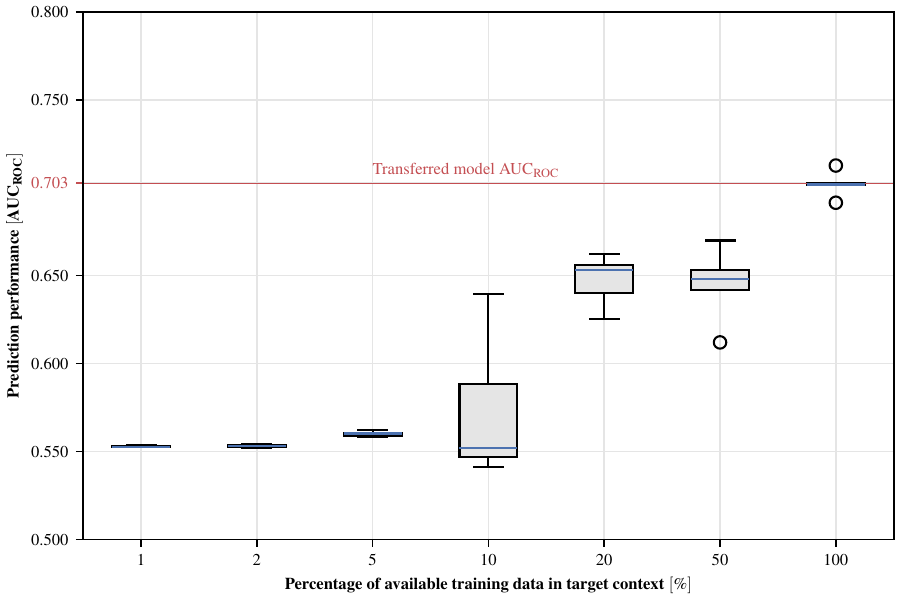}
	\caption{Prediction performance of transferred models compared to models trained on increasing training event data in the target context using the pre-trained embedding model \emph{all-mpnet-base-v2} for the intra-organizational use case over five runs.}
 \label{fig:intracaseScale} 
\end{figure}

To provide an overview of how much target data should be available to train a model outperforming our transferred model (not fine-tuned), Figure~\ref{fig:intracaseScale} shows the $\text{AUC}_{\text{ROC}}$ for training a model solely on event data available in the target context with the first 1\%, 2\%, 5\%, 10\%, 20\%, 50\%, and 100\% of traces using the \emph{all-mpnet-base-v2} embedding model (see Section~\ref{sec:results:intra_actemb}) for activity encoding and the relative cross-domain approach (see Section~\ref{sec:results:intra_timenc}) for timestamp encoding. 

In terms of $\text{AUC}_{\text{ROC}}$, the transferred model achieves a higher $\text{AUC}_{\text{ROC}}$ compared to the majority of the models trained on any amount of available event data of the target context. The prediction performance of the models gradually improves with more training event data.
However, our proposed \gls{tl}-based technique achieves a comparable prediction performance in terms of $\text{AUC}_{\text{ROC}}$ to models trained on the full training event data.
These results show that our transferred model can be used in the intra-organizational use case as a considerable substitute for training a separate model in the target context.

\subsection{Use case II: Inter-organizational transfer}

For the inter-organizational use case, we proceeded as for the intra-organizational use case and evaluated the prediction performance of our technique compared to traditional \gls{ppm} approaches, different activity embeddings, as well as embedding quality and time encodings. Furthermore, we examined the transfer relevance by determining the quantity of training event data required in the target context to attain a prediction performance comparable to that of our proposed technique.

\subsubsection{Prediction performance for transfer learning-based compared to traditional predictive process monitoring}
\label{sec:results:inter_trad}

Table~\ref{tab:Results_inter} shows the prediction performance of our proposed \gls{tl}-based \gls{ppm} technique and the two baseline approaches, compared to five traditional \gls{ppm} techniques, for the inter-organizational use case. We chose the best activity encoding (see Section~\ref{sec:results:inter_actemb}) and timestamp encoding (see Section~\ref{sec:results:inter_timenc}) for our proposed technique.
Similar to the intra-organizational use case, our proposed \gls{tl}-based \gls{ppm} technique outperforms all traditional \gls{ppm} approaches in recall, F1 score, and $\text{AUC}_{\text{ROC}}$.

The traditional \gls{ppm} technique that employs the \Gls{xgb} model attains a higher recall, whereas the approach with the random forest model yields a superior \gls{mcc}.
Comparing the results of the transferred models with the baseline approach \emph{Train and test on $L_T$}, the transferred models come close to the prediction performance of the baseline approach \emph{Train and test on $L_T$} in most metrics.

\begin{table}[ht!]
\centering
\caption{Prediction performance of our proposed \gls{tl}-based \gls{ppm} technique and traditional \gls{ppm} approaches for the inter-organizational use case (average and standard deviation over five runs).}
\label{tab:Results_inter}
\resizebox{\textwidth}{!}{ 
\begin{tabular}{@{}lrrrrr@{}}
\toprule
\textbf{}                                       \textbf{ML approach}      & \textbf{Precision} & \textbf{Recall} & \textbf{F1-score} & \textbf{MCC} &\textbf{$\text{AUC}_{\text{ROC}}$} 
\\ \midrule

\multicolumn{6}{l}{\textsc{TL-based PPM Technique}}\\
\textbf{Our proposed technique}                       &                   &                   &                   &              &    \\
Train and test on $L_S$    & 0.735 ($\pm$.003)   & 0.694 ($\pm$.003)     &  0.708 ($\pm$.002)  &  0.293 ($\pm$.007) &  0.730 ($\pm$.005)\\
Train and test on $L_T$    & 0.769 ($\pm$.012) &  0.769 ($\pm$.037)     &  0.766 ($\pm$.024)  & 0.330 ($\pm$.034) & 0.775 ($\pm$.018)\\
Train on $L_S$, test on $L_T$ & 0.739 ($\pm$.020)&  \textbf{0.755} ($\pm$.050) &   \textbf{0.724} ($\pm$.021) &   0.206 ($\pm$.029) & \textbf{0.711} ($\pm$.020) \\

\midrule 

\multicolumn{6}{l}{\textsc{Traditional PPM Techniques}}\\
\textbf{LSTM} & & & & & \\
Train and test on $L_S$    & 0.737  ($\pm$.001)    &   0.678  ($\pm$.004)  &  0.695  ($\pm$.004) & 0.292  ($\pm$.004) & 0.721  ($\pm$.005) \\
Train and test on $L_T$           &  0.676  ($\pm$.047) & 0.775  ($\pm$.000)  & 0.678  ($\pm$.001) & 0.011  ($\pm$.016) & 0.567  ($\pm$.002) \\
Train on $L_S$, test on $L_T$    &    0.050 ($\pm$.000)       &  0.224  ($\pm$.000)         &   0.082  ($\pm$.000)        &   -0.029 ($\pm$.000)  & 0.435 ($\pm$.012) \\

\textbf{Logistic Regression}                         &                   &                   &                   &              &    \\
Train and test on $L_S$    & 0.708 ($\pm$.000)   & 0.740 ($\pm$.000)     & 0.714 ($\pm$.000)  & 0.218 ($\pm$.000) & 0.728 ($\pm$.000)\\
Train and test on $L_T$    & 0.759 ($\pm$.000)   & 0.790 ($\pm$.000)     & 0.754 ($\pm$.000)  & 0.269 ($\pm$.000) & 0.670 ($\pm$.000)\\
Train on $L_S$, test on $L_T$ & 0.652 ($\pm$.000) & 0.286 ($\pm$.000)     & 0.231 ($\pm$.000)  & 0.000 ($\pm$.000)   & 0.486 ($\pm$.000)\\

\textbf{XGBoost}                      &                   &                   &                   &              &    \\
Train and test on $L_S$    & 0.714 ($\pm$.002)   & 0.743 ($\pm$.002)     & 0.720 ($\pm$.001)  & 0.234 ($\pm$.004) & 0.738 ($\pm$.002)\\
Train and test on $L_T$    & 0.793 ($\pm$.000)   & 0.811 ($\pm$.000)     & 0.793 ($\pm$.000)  & 0.386 ($\pm$.000) & 0.759 ($\pm$.000)\\
Train on $L_S$, test on $L_T$ & \textbf{0.828} ($\pm$.003) & 0.257 ($\pm$.044)     & 0.143 ($\pm$.080)  & 0.086 ($\pm$.059) & 0.604 ($\pm$.004)\\

\textbf{Random Forest}                      &                   &                   &                   &              &    \\
Train and test on $L_S$    & 0.703 ($\pm$.001)   & 0.710 ($\pm$.001)     & 0.706 ($\pm$.001)  & 0.218 ($\pm$.002) & 0.692 ($\pm$.001)\\
Train and test on $L_T$    & 0.752 ($\pm$.001)   & 0.756 ($\pm$.002)     & 0.754 ($\pm$.001)  & 0.286 ($\pm$.003) & 0.724 ($\pm$.002)\\
Train on $L_S$, test on $L_T$ & 0.827 ($\pm$.002) & 0.385 ($\pm$.002)     & 0.363 ($\pm$.004)  & \textbf{0.227} ($\pm$.001) & 0.596 ($\pm$.010)\\

\textbf{Decision Tree}                      &                   &                   &                   &              &    \\
Train and test on $L_S$    & 0.684 ($\pm$.001)   & 0.681 ($\pm$.001)     &  0.683 ($\pm$.001)  &  0.169 ($\pm$.001) &  0.590 ($\pm$.001)\\
Train and test on $L_T$    & 0.738 ($\pm$.001)   & 0.726 ($\pm$.001)     &  0.732 ($\pm$.001)  &  0.245 ($\pm$.004) &  0.676 ($\pm$.003)\\
Train on $L_S$, test on $L_T$ & 0.794 ($\pm$.002) & 0.402 ($\pm$.006)     & 0.393 ($\pm$.008)   & 0.204 ($\pm$.006)  & 0.597 ($\pm$.004)\\

\bottomrule
\multicolumn{6}{p{18cm}}{\textit{Note.} Best results for training on $L_S$ and testing on $L_T$ are marked in bold.} 
\end{tabular}
}
\end{table}

\subsubsection{Prediction performance for different activity embedding models}
\label{sec:results:inter_actemb}

In terms of the activity embedding, we again examined seven different embedding models, belonging to the categories static word, contextual word, and contextual embedding models. Table~\ref{tab:Results_inter_actemb} shows the results of these seven embeddings. For the inter-organizational use case, the static word embedding model \emph{glove-wiki-gigaword-100} achieves the highest recall, F1-score, and $\text{AUC}_{\text{ROC}}$. The contextual word embedding model \emph{bert-base-uncased} outperforms other embedding models in terms of precision and \gls{mcc}.

\begin{table}[ht!]
\centering
\caption{Prediction performance of our \gls{tl}-based \gls{ppm} technique with the proposed activity encoding created with different pre-trained embedding models for the inter-organizational use case (average and standard deviation over five runs).}
\label{tab:Results_inter_actemb}
\resizebox{\textwidth}{!}{ 
\begin{tabular}{@{}lrrrrr@{}}
\toprule
\textbf{}                                        \textbf{Embedding model}     & \textbf{Precision} & \textbf{Recall} & \textbf{F1-score} & \textbf{MCC} &\textbf{$\text{AUC}_{\text{ROC}}$}
\\ \midrule

\multicolumn{3}{l}{\textsc{Static Word Embedding Models}}\\

\textbf{glove-wiki-gigaword-100}                       &                  &                   &                   &              &    \\
Train and test on $L_S$    & 0.735 ($\pm$.003)   & 0.694 ($\pm$.003)     &  0.708 ($\pm$.002)  &  0.293 ($\pm$.007) &  0.730 ($\pm$.005)\\
Train and test on $L_T$    & 0.769 ($\pm$.012) &  0.769 ($\pm$.037)     &  0.766 ($\pm$.024)  & 0.330 ($\pm$.034) & 0.775 ($\pm$.018)\\
Train on $L_S$, test on $L_T$ & 0.739 ($\pm$.020)&  \textbf{0.755} ($\pm$.050) &   \textbf{0.724} ($\pm$.021) &   0.206 ($\pm$.029) & \textbf{0.711} ($\pm$.020) \\

\textbf{glove-wiki-gigaword-300}                         &                   &                   &                   &              &    \\
Train and test on $L_S$    & 0.738 ($\pm$.003)   & 0.693 ($\pm$.009)     & 0.708 ($\pm$.006)  & 0.300 ($\pm$.005)   & 0.733 ($\pm$.003)\\
Train and test on $L_T$    & 0.756 ($\pm$.025)   & 0.745 ($\pm$.062)     & 0.747 ($\pm$.047)  & 0.294 ($\pm$.074) & 0.748 ($\pm$.037)\\
Train on $L_S$, test on $L_T$ & 0.723 ($\pm$.021) & 0.669 ($\pm$.054)     & 0.685 ($\pm$.040)  & 0.193 ($\pm$.052) & 0.689 ($\pm$.043)\\

\textbf{word2vec-google-news-300}                      &                   &                   &                   &              &    \\
Train and test on $L_S$         & 0.734 ($\pm$.005) & 0.701 ($\pm$.010) & 0.713 ($\pm$.006) & 0.294 ($\pm$.008) & 0.732 ($\pm$.004) \\
Train and test on $L_T$         & 0.769 ($\pm$.005) & 0.774 ($\pm$.014) & 0.770 ($\pm$.009) & 0.333 ($\pm$.012) & 0.784 ($\pm$.005) \\
Train on $L_S$, test on $L_T$   & 0.735 ($\pm$.030) & 0.718 ($\pm$.070) & 0.713 ($\pm$.041) & 0.215 ($\pm$.073) & 0.689 ($\pm$.021) \\
 \midrule

\multicolumn{3}{l}{\textsc{Contextual Word Embedding Models}}\\

\textbf{bert-base-cased}                      &                   &                   &                   &              &    \\
Train and test on $L_S$           & 0.738 ($\pm$.002) & 0.691 ($\pm$.010) & 0.707 ($\pm$.008) & 0.300 ($\pm$.003) & 0.735 ($\pm$.002)\\
Train and test on $L_T$           & 0.676 ($\pm$.042) & 0.735 ($\pm$.045) & 0.692 ($\pm$.013) & 0.084 ($\pm$.050) & 0.638 ($\pm$.007)\\
Train on $L_S$, test on $L_T$     & 0.760 ($\pm$.057) & 0.484 ($\pm$.138) & 0.483 ($\pm$.142) & 0.173 ($\pm$.049) & 0.646 ($\pm$.019)\\

\textbf{bert-base-uncased}                      &                   &                   &                   &              &    \\
Train and test on $L_S$    & 0.735 ($\pm$.004)   & 0.700 ($\pm$.011)     &  0.713 ($\pm$.008)  &  0.296 ($\pm$.009)&  0.736 ($\pm$.006)\\
Train and test on $L_T$           & 0.715 ($\pm$.062) &  0.766 ($\pm$.038)     &  0.727 ($\pm$.052)        & 0.174 ($\pm$.174) & 0.690 ($\pm$.064)\\
Train on $L_S$, test on $L_T$    & \textbf{0.766} ($\pm$.043)&  0.618 ($\pm$.173) &   0.619 ($\pm$.167) &   \textbf{0.245} ($\pm$.073)       & 0.692 ($\pm$.042) \\
\midrule

\multicolumn{3}{l}{\textsc{Contextual Sentence Embedding Models}}\\

\textbf{all-MiniLM-L12-v2}                      &                   &                   &                   &              &    \\
Train and test on $L_S$    & 0.736 ($\pm$.003)   & 0.694 ($\pm$.009)     &  0.708 ($\pm$.007)  &  0.296 ($\pm$.006) &  0.732 ($\pm$.006)\\
Train and test on $L_T$    & 0.772 ($\pm$.007)   & 0.781 ($\pm$.016)     &  0.775 ($\pm$.009)  &  0.342 ($\pm$.016) &  0.782 ($\pm$.005)\\
Train on $L_S$, test on $L_T$ & 0.721 ($\pm$.014) & 0.668 ($\pm$.089)     &  0.681 ($\pm$.069)  &  0.188 ($\pm$.051) &  0.681 ($\pm$.026)\\

\textbf{all-mpnet-base-v2}                      &                   &                   &                   &              &    \\
Train and test on $L_S$    & 0.736 ($\pm$.003) & 0.691 ($\pm$.003) & 0.706 ($\pm$.002) & 0.294 ($\pm$.006) & 0.728 ($\pm$.005)\\
Train and test on $L_T$    & 0.766 ($\pm$.012) & 0.775 ($\pm$.019) & 0.769 ($\pm$.014) & 0.324 ($\pm$.031) & 0.762 ($\pm$.022)\\
Train on $L_S$, test on $L_T$    & 0.707 ($\pm$.009) & 0.550 ($\pm$.040) & 0.585 ($\pm$.039) & 0.130 ($\pm$.017) & 0.655 ($\pm$.014)\\

\bottomrule
\multicolumn{6}{p{18cm}}{\textit{Note.} Best results for training on $L_S$ and testing on $L_T$ are marked in bold.}

\end{tabular}
}
\end{table}

\subsubsection{Activity embedding quality}

\begin{figure}[ht!]
\centering
\includegraphics[width=0.8\textwidth]{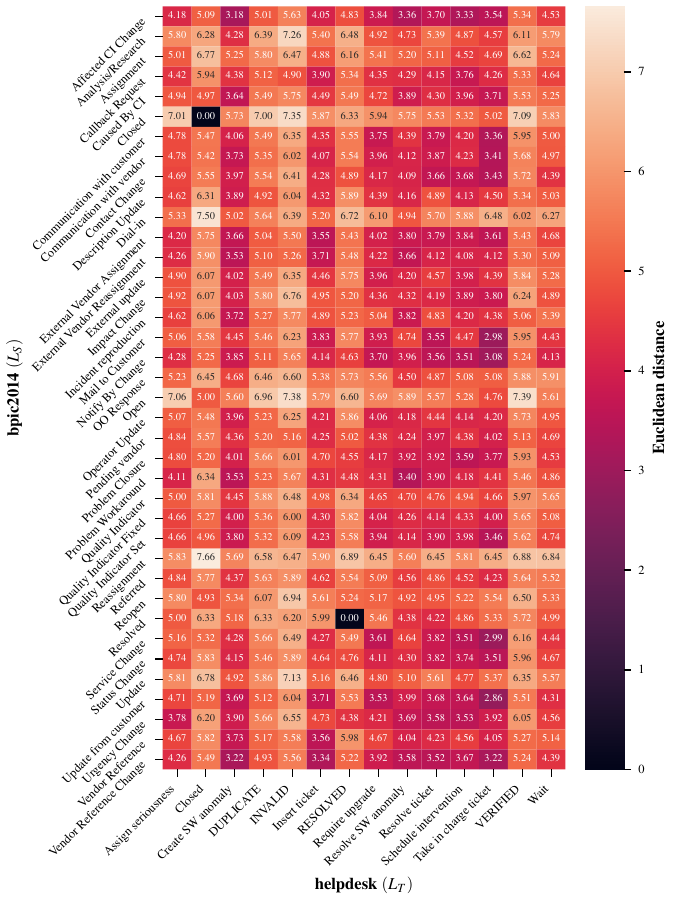}
\caption{Difference in embedding with model \emph{glove-wiki-gigaword-100} of activities in $L_S$ and $L_T$ for the inter-organizational use case.}
\label{fig:actem_inter} 
\end{figure}

Figure~\ref{fig:actem_inter} shows the Euclidean distance between the embedding vectors of the activities from $L_S$ and $L_T$, created via the \emph{glove-wiki-gigaword-100} embedding model.\footnote{The figures showing the Euclidean distance between activity embedding vectors created with the other embedding models of our setting can be found in the online repository of this work.}

The source and target logs share only two common activities, \emph{Closed} and \emph{RESOLVED}, which are represented by an activity embedding distance of 0.00. 
The other twelve activities in $L_T$ cannot be found in the source log $L_S$. Interestingly, there is no clear match based on the smallest Euclidean distance between the embedding vectors for any of the 12 activities in $L_T$ with any of the activities in $L_S$.
However, based on the embedding distance, these twelve activities can be categorized into two groups. The first group contains seven activities, which have an overall closer match to other activities in $L_S$. These are \emph{Create SW anomaly}, \emph{Insert ticket}, \emph{Require upgrade}, \emph{Resolve SW anomaly}, \emph{Resolve ticket}, \emph{Schedule intervention}, and \emph{Take in charge ticket}. 
The other five activities in $L_T$, \emph{Assign seriousness}, \emph{DUPLICATE}, \emph{INVALID}, \emph{VERIFIED}, and \emph{Wait}, are overall further apart from activities in $L_S$ based on the Euclidean distance between the embedding vectors. However, the activities written in capital letters are outliers in terms of their frequency occurring in $L_T$ (see Table~\ref{tab:actslogs_inter} in Section~\ref{sec:eval:usecaseII}).

\subsubsection{Prediction performance for different timestamp encoding approaches}
\label{sec:results:inter_timenc}

Similar to the intra-organizational use case, we also examined five additional encoding strategies for the time-based feature \emph{duration since start}, belonging to the categories time-based encoding and autoencoder-based encoding, and compared them to our technique's relative cross-domain encoding. 
The results in Table~\ref{tab:Results_intra_timenc} indicate a clear trend towards the relative cross-domain encoding as it achieves the highest precision, recall, F1-score, \gls{mcc}, and $\text{AUC}_{\text{ROC}}$ out of the six encoding strategies. The prediction performance of our technique using the time-based and autoencoder-based encoding strategies is comparably lower.

\begin{table}[ht!]
\centering
\caption{Prediction performance of our \gls{tl}-based \gls{ppm} technique with the proposed cross-domain encoding of the \emph{duration since start} time feature and other encoding approaches for the inter-organizational use case (average and standard deviation over five runs).}
\label{tab:Results_inter_timenc}
\resizebox{\textwidth}{!}{ 
\begin{tabular}{@{}lrrrrr@{}}
\toprule
\textbf{}                                        \textbf{Encoding approach}     & \textbf{Precision} & \textbf{Recall} & \textbf{F1-score} &\textbf{MCC}   & \textbf{$\text{AUC}_{\text{ROC}}$}
\\ \midrule

\multicolumn{6}{l}{\textsc{Relative Cross-domain Encoding}}\\
Train and test on $L_S$    & 0.735 ($\pm$.003)   & 0.694 ($\pm$.003)     &  0.708 ($\pm$.002)  &  0.293 ($\pm$.007) &  0.730 ($\pm$.005)\\
Train and test on $L_T$    & 0.769 ($\pm$.012) &  0.769 ($\pm$.037)     &  0.766 ($\pm$.024)  & 0.330 ($\pm$.034) & 0.775 ($\pm$.018)\\
Train on $L_S$, test on $L_T$ & \textbf{0.739} ($\pm$.020)&  \textbf{0.755} ($\pm$.050) &   \textbf{0.724} ($\pm$.021) &   \textbf{0.206} ($\pm$.029) & \textbf{0.711} ($\pm$.020) \\ 
\midrule
\multicolumn{6}{l}{\textsc{Time-based Encoding}}\\
\textbf{Hour}                       &                   &                   &                   &              &    \\
Train and test on $L_S$    &  0.740 ($\pm$.002)    &   0.678 ($\pm$.012)  &  0.696 ($\pm$.010) & 0.299 ($\pm$.004)  & 0.728 ($\pm$.003) \\
Train and test on $L_T$           &  0.686 ($\pm$.018) & 0.726 ($\pm$.050)  & 0.687 ($\pm$.007) & 0.078 ($\pm$.045) & 0.633 ($\pm$.007) \\
Train on $L_S$, test on $L_T$    &    0.702 ($\pm$.037)       &  0.657 ($\pm$.126)         &   0.637 ($\pm$.095)        &   0.093 ($\pm$.050)   & 0.609 ($\pm$.013)  \\ 
\textbf{Weekday}                       &                   &                   &                   &              &    \\
Train and test on $L_S$    &  0.739 ($\pm$.004)    &   0.687 ($\pm$.013)   &  0.704 ($\pm$.010) & 0.300 ($\pm$.009) & 0.736 ($\pm$.001) \\
Train and test on $L_T$           &  0.686 ($\pm$.013) & 0.760 ($\pm$.006)  & 0.697 ($\pm$.006) & 0.072 ($\pm$.026) & 0.636 ($\pm$.005) \\
Train on $L_S$, test on $L_T$    &    0.687 ($\pm$.001)       &  0.724 ($\pm$.026)         &   0.698 ($\pm$.004)        &   0.094 ($\pm$.009)   & 0.624 ($\pm$.009) \\ 
\textbf{Month}                       &                   &                   &                   &              &    \\
Train and test on $L_S$    &  0.732 ($\pm$.005)    &   0.689 ($\pm$.005)  &  0.703 ($\pm$.003) & 0.284 ($\pm$.009) & 0.724 ($\pm$.005) \\
Train and test on $L_T$           &  0.680 ($\pm$.012) & 0.719 ($\pm$.034)  & 0.690 ($\pm$.005) & 0.074 ($\pm$.037) & 0.631 ($\pm$.002)  \\
Train on $L_S$, test on $L_T$    &    0.687 ($\pm$.005)      &  0.674 ($\pm$.047)         &   0.676 ($\pm$.023)        &   0.096 ($\pm$.014)  & 0.628 ($\pm$.005) \\ 
\textbf{Hour + Weekday + Month}                       &                   &                   &                   &              &    \\
Train and test on $L_S$    &  0.729  ($\pm$.003)    &   0.707  ($\pm$.016)  &  0.715  ($\pm$.010) & 0.282  ($\pm$.006) & 0.724  ($\pm$.005) \\
Train and test on $L_T$           &  0.683  ($\pm$.013) & 0.737  ($\pm$.031)  & 0.694  ($\pm$.005) & 0.073  ($\pm$.030) & 0.628  ($\pm$.010) \\
Train on $L_S$, test on $L_T$    &    0.681  ($\pm$.002)       &  0.678  ($\pm$.048)         &   0.675  ($\pm$.024)        &   0.080  ($\pm$.007)  & 0.613 ($\pm$.008) \\ 
\midrule

\multicolumn{3}{l}{\textsc{Autoencoder-based Encoding}}\\
Train and test on $L_S$           & 0.740 ($\pm$.001) & 0.684 ($\pm$.007) & 0.701 ($\pm$.006) & 0.301 ($\pm$.002) & 0.732 ($\pm$.004) \\
Train and test on $L_T$           & 0.786 ($\pm$.013) & 0.801 ($\pm$.016) & 0.787 ($\pm$.010) & 0.372 ($\pm$.027) & 0.785 ($\pm$.006) \\
Train on $L_S$, test on $L_T$  & 0.732 ($\pm$.058) & 0.474 ($\pm$.148) & 0.470 ($\pm$.177) & 0.135 ($\pm$.082) & 0.627 ($\pm$.029) \\

\bottomrule
\multicolumn{6}{p{18cm}}{\textit{Note.} Best results for training on $L_S$ and testing on $L_T$ are marked in bold.}

\end{tabular}
}
\end{table}

\subsubsection{Transfer relevance}

\begin{figure}[ht] 
\centering
\includegraphics[width=0.8\textwidth]{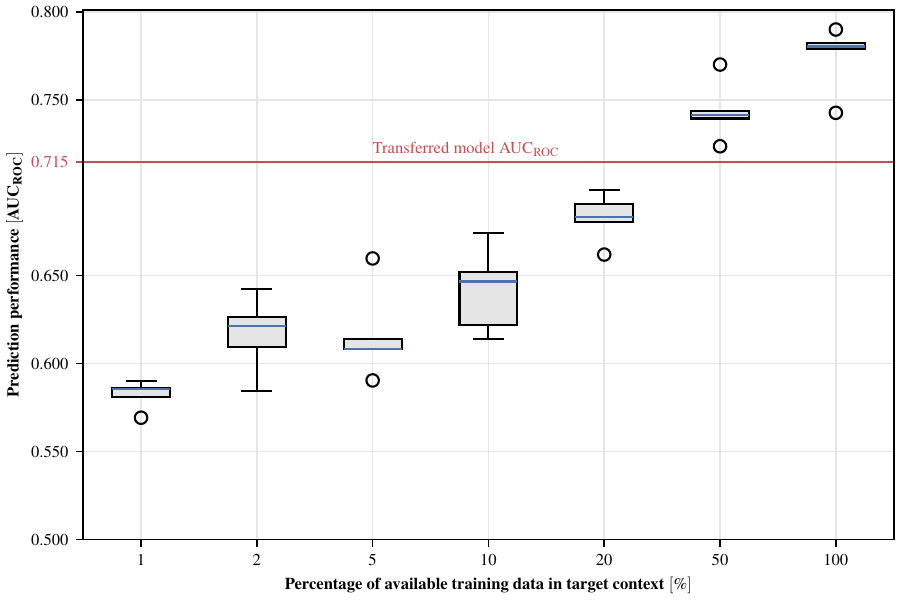}
	\caption{Prediction performance of transferred models compared to models trained on increasing training event data in the target context using the pre-trained embedding model \emph{glove-wiki-gigaword-100} for the inter-organizational use case over five runs.}
 \label{fig:intercaseScale} 
\end{figure}

To provide an overview of how much event data of the target context should be available to train a model outperforming our transferred models (not fine-tuned), Figure~\ref{fig:intercaseScale} shows the $\text{AUC}_{\text{ROC}}$ for training a model solely on event data available in the target context with the first 1\%, 2\%, 5\%, 10\%, 20\%, 50\%, and 100\% of traces using the \emph{glove-wiki-gigaword-100} embedding model (see Section~\ref{sec:results:inter_actemb}) for activity encoding and the relative cross-domain approach (see Section~\ref{sec:results:inter_timenc}) for time encoding.

In terms of $\text{AUC}_{\text{ROC}}$, the transferred model achieves a higher prediction performance compared to the majority of the models trained on 1\%, 2\%, 5\%, 10\%, 20\% of the available event data of the target context. 
All models with more than 50\% of the event data of the target context outperform our transferred model in terms of $\text{AUC}_{\text{ROC}}$.
These results show that our transferred model can be used in the inter-organizational use case as a starting point if event data to train a model in the target context are not readily available in sufficient amounts.

\section{Discussion}
\label{sec:discussion}

\subsection{Theoretical implications}

Our study has multiple implications for research.
%
First, the evaluation results show that the proposed \gls{tl}-based \gls{ppm} technique clearly outperforms all the traditional \gls{ppm} techniques considered in the intra- and inter-organizational use cases in terms of prediction performance. This suggests that a similar understanding of both the source and target contexts is essential for the effective transfer of process knowledge. This is even more important when there is no or very little event data available in the target context for fine-tuning a transferred model. Our work proposes two approaches for aligning the domains of the contexts: i) pre-trained embedding models for activity encoding and ii) a relative cross-domain mapping for timestamp encoding. 

Second, our results in the inter- and intra-organizational use cases indicate that pre-trained embedding models can be a better choice to encode the activities for \gls{tl} in \gls{ppm} than using a sparse encoding technique, such as one-hot encoding. 
In addition, depending on the domain (event data) in the source and target context, different pre-trained embedding models lead to the best prediction performance of the transferred model in the target context. 
One possible explanation for this is that there is a correlation between the data used to pre-train the embedding models and the activity attribute values in the event data of the source and target contexts. 
%

Third, we show how embedding models can facilitate \gls{tl} in \gls{ppm}. 
These models enable the encoding of process activities from different business processes of the same type that have different names but similar semantic meanings. 
This is possible with embedding models because they are trained in a self-supervised manner with auxiliary tasks (e.g., predicting a word in a sentence given the remaining words) to learn the representation of words while capturing their semantics.  

Moreover, in our experiments, we observed that two important factors for the transfer of knowledge from one process to another are i) the prediction target and ii) the process information considered for the transfer. 
Regarding the prediction target, a process-agnostic binary outcome prediction target, like the in-time prediction, is easier for the transfer than a process-specific outcome prediction or the next activity prediction. This is because the latter prediction targets differ more conceptually in the source and target contexts. 
Concerning the considered process information, a transfer that relies solely on the control flow is typically simpler than a transfer that relies on the control flow and additional time or data attributes. This is because the consideration of each additional attribute in the event data of the source and target contexts, with its values and distribution, increases the complexity. Nevertheless, to show that information beyond the control flow can be transferred from source to target contexts, we additionally considered the timestamp attribute in the intra- and inter-organization use cases of our evaluation and extracted the time feature \emph{duration since start} as part of the relative cross-domain encoding.

\subsection{Practical implications}
Our work also has implications for practice.
With a \gls{tl}-based \gls{ppm} technique like ours, the adoption of \gls{ppm} can be encouraged in practice. This is specifically beneficial for organizations or areas in organizations without suitable resources, such as computational power, skills or training of employees, or event data for model training. 
These organizations can use the base model and further resources from another organization that already uses \gls{ppm} successfully for similar business processes. This enables them to predict, for example, process outcomes or throughput times, and to use these predictions to mitigate risks and decrease costs. 
 
Furthermore, the proposed technique is scalable to incorporate the event data of multiple processes of the same type in the source and target contexts from a conceptual perspective. 
In this research, we used one event log as a source domain to train our base model, and one event log as a target domain to apply the base model.
However, it is conceivable that multiple event logs can be used as a domain in the source context for training our base model. Similarly, multiple event logs can be used as different domains in the target context to which our base model is applied. 
To scale our technique in such a way, concepts from the field of federated learning can be incorporated \citep{Verman2019}.
As our technique uses an embedding model for activity encoding, scaling is easier than, for example, with a fixed encoding strategy like one-hot encoding, where each activity included in the event data of the source and target contexts has a specific position in the created feature vectors.

\subsection{Future research} 

There are four promising directions for future research.  
First, fine-tuning of the embedding models used in our technique for feature-based \gls{tl} can be investigated. We suspect that this will improve the prediction performance of the model used in the target context. As the source context can comprise multiple business processes and their event data, certain strategies for fine-tuning can be realized to adapt the embedding models to multiple similar processes in the same context in a controlled way. For example, such a strategy could be to randomly select 1,000 traces from each business process's event log to ensure a fair fine-tuning across multiple business processes with varying amounts of event data.    
Second, the transfer of information about additional event log dimensions can be investigated. In this paper, we have considered information on the control flow and the time perspectives in the design of our technique. However, information on additional attributes available in the event data, such as data attributes, might be valuable for the transfer in \gls{ppm}. 
For this purpose, approaches from the research stream of multidimensional process representation creation and learning in \gls{bpm} \citep{weinzierl2024machine} can be helpful to find a representation of multiple event log attributes, facilitating the process knowledge transfer from source to target contexts. 
%
The question of which information is necessary for the knowledge transfer is closely connected to the information used and its representation.  
This question can be investigated with explainable artificial intelligence (XAI) approaches, which are common in \gls{ppm}~\citep{2021ecisxpbpm}.  

Third, the application of \gls{tl} in the context of different types of real business processes can be investigated. In the case of those business processes, the complexity of \gls{tl} is further increased because they differ more structurally and semantically. 
For example, most of the activities, as well as additional data attributes in the event data of a business process in the source context, can differ from the event data of a business process in the target context.  

Fourth, the building of process-specific foundation models that are universally usable for various data-driven \gls{bpm} tasks along the \gls{bpm} lifecycle \citep{rizk2023case,buss2024processllm} is another promising avenue for future research. For this, existing foundation models for (multivariate) time series can serve as a starting point for model development~\citep[e.g.,][]{ye2024survey}, while available benchmark event logs and other process-relevant resources (e.g., fundamental \gls{bpm} books, scientific \gls{bpm} papers, or additional descriptions of event logs) can be used as a data basis for initial model training. As foundational models can support multiple modalities, types of process data going beyond text-transformed event data (e.g., process model collections or process-attached images, audio tracks, or video sequences) are additionally conceivable for model fine-tuning. In order to fine-tune a foundation model with event data from various organizations, the concept of federated learning can be essential to overcome data privacy restrictions~\citep{Verman2019}. 

\section{Conclusion}
\label{sec:conclusion}

In this paper, we proposed a \gls{tl}-based technique for \gls{ppm}, allowing organizations without suitable event data or other relevant resources to create predictions in running business processes for effective decision support. 
For this purpose, we adopted the computational design science paradigm \citep{rai2017editor} and designed an artifact in the form of a \gls{tl}-based technique for \gls{ppm}, consisting of three phases: i) \textit{initial model building on source}, ii) \textit{transfer base model from source to target}, and iii) \textit{online application of model to target}. 
Based on the instantiation of our technique in both a real-life intra- and an inter-organizational use case, including event logs for \gls{itsm} processes from two organizations, we evaluated its prediction performance. 
Our results suggest that knowledge of one business process can be transferred to a similar business process in the same or a different organization to enable effective \gls{ppm} in the target context.

\section*{Acknowledgments}
Martin Matzner acknowledges funding from the Bavarian State Ministry of Economic Affairs, Regional Development and Energy (StMWi) on \textquote{PräMi} (Grant DIK-2307-0002// DIK0533/01) and the German Research Foundation (DFG) on \textquote{CoPPA} (Grant 456415646). Furthermore, the authors want to thank Matthias Stierle for his contribution to the initial discussions on the general idea of this research project and Florian Gatzlaff for his support during the project's early stages.


\setstretch{0.5}
\bibliographystyle{spbasic-bise}
\bibliography{references}

\setstretch{1.0}
\clearpage
\appendix
\gdef\thesection{\Alph{section}} 
\makeatletter
\renewcommand\@seccntformat[1]{Appendix \csname the#1\endcsname.\hspace{0.5em}}
\makeatother

\setcounter{page}{1}

\vspace{1cm}
\begin{center}
\Large Online Appendix
\end{center}
\vspace{0.5cm}

\section{Prediction performance for complete model compared to model sub-variants}
\label{app:SourceLogScaling}

We conducted an additional evaluation in the form of an ablation analysis to assess the importance of two main components of our proposed model on prediction performance. These components include the second \gls{lstm} layer, which transforms the model into an encoder, and the dedicated parameter initialization, which is performed before model training. When the dedicated parameter initialization is deactivated, all model weights and biases are initialized uniformly. Both model components are designed to improve learning from the event data of the source context and facilitate the transfer of process knowledge to the target context.  

Table~\ref{tab:ResultsApproachComponents} shows that the complete model of our proposed \gls{tl}-based technique clearly outperforms its sub-variants in terms of prediction performance when it is trained on $L_S$ and tested on $L_T$ for the inter-organizational use case. This confirms the importance of using both components in the prediction model for an effective process knowledge transfer from source to target contexts.     

\begin{table}[ht!]
\centering
\caption{Prediction performance of complete model compared to its sub-variants for the inter-organizational use case (average and standard deviation over five runs).}
\label{tab:ResultsApproachComponents}
\resizebox{\textwidth}{!}{ 
\begin{tabular}{@{}lrrrrr@{}}
\toprule
\textbf{}                                       \textbf{Scenario}      & \textbf{Precision} & \textbf{Recall} & \textbf{F1-score} & \textbf{MCC} &\textbf{$\text{AUC}_{\text{ROC}}$} 
\\ \midrule

\multicolumn{6}{l}{\textsc{Complete Model}}\\
\multicolumn{6}{@{}l}{\textbf{Model with two LSTM layers and dedicated parameter initialization}}\\

Train and test on $L_S$    & 0.735 ($\pm$.003)   & 0.694 ($\pm$.003)     &  0.708 ($\pm$.002)  &  0.293 ($\pm$.007) &  0.730 ($\pm$.005)\\
Train and test on $L_T$    & 0.769 ($\pm$.012) &  0.769 ($\pm$.037)     &  0.766 ($\pm$.024)  & 0.330 ($\pm$.034) & 0.775 ($\pm$.018)\\
Train on $L_S$, test on $L_T$ & \textbf{0.739} ($\pm$.020)&  \textbf{0.755} ($\pm$.050) &   \textbf{0.724} ($\pm$.021) &   \textbf{0.206} ($\pm$.029) & \textbf{0.711} ($\pm$.020) \\
\midrule

\multicolumn{6}{l}{\textsc{Sub-variants of Model}}\\
\multicolumn{6}{@{}l}{\textbf{Model with two LSTM layers and without dedicated parameter initialization}}\\

Train and test on $L_S$    & 0.737 ($\pm$.005)   & 0.691 ($\pm$.009)     &  0.706 ($\pm$.006)  &  0.296 ($\pm$.010)&  0.730 ($\pm$.005)\\
Train and test on $L_T$    & 0.770 ($\pm$.011) &  0.783 ($\pm$.015)     &  0.775 ($\pm$.011)  & 0.336 ($\pm$.027) & 0.767 ($\pm$.015)\\
Train on $L_S$, test on $L_T$    & 0.712 ($\pm$.037)&  0.701 ($\pm$.048) &   0.702 ($\pm$.035) &   0.167 ($\pm$.099) & 0.666 ($\pm$.063) \\

\multicolumn{6}{@{}l}{\textbf{Model with one LSTM layer and with dedicated parameter initialization}}\\

Train and test on $L_S$          & 0.738 ($\pm$.004)   & 0.696 ($\pm$.009)     &  0.710 ($\pm$.007)  &  0.301 ($\pm$.007) &  0.734 ($\pm$.004)\\
Train and test on $L_T$          & 0.756 ($\pm$.021)   & 0.743 ($\pm$.058)     &  0.745 ($\pm$.043)  &  0.292 ($\pm$.062) &  0.728 ($\pm$.030)\\
Train on $L_S$, test on $L_T$    & 0.719 ($\pm$.023)   & 0.694 ($\pm$.046)     &  0.693 ($\pm$.009)  &  0.171 ($\pm$.058) &  0.701 ($\pm$.022)\\

\multicolumn{6}{@{}l}{\textbf{Model with one LSTM layer and without dedicated parameter initialization}}\\
Train and test on $L_S$    & 0.739 ($\pm$.005)   & 0.686 ($\pm$.008)     &  0.702 ($\pm$.006)  &  0.299 ($\pm$.009)&  0.731 ($\pm$.006)\\
Train and test on $L_T$           & 0.764 ($\pm$.016) &  0.774 ($\pm$.025)     &  0.767 ($\pm$.019)        & 0.318 ($\pm$.045) & 0.741 ($\pm$.032)\\
Train on $L_S$, test on $L_T$    & 0.702 ($\pm$.026)&  0.688 ($\pm$.040) &   0.687 ($\pm$.017) &   0.133 ($\pm$.068)       & 0.670 ($\pm$.044) \\

\bottomrule
\multicolumn{6}{p{18cm}}{\textit{Note.} Best results for training on $L_S$ and testing on $L_T$ are marked in bold.} 
\end{tabular}
}
\end{table}

\newpage
\section{Prediction performance for different prefix encoding approaches}
\label{app:PrefixEncodingTechniques}

We conducted an additional evaluation to assess the effect of index-based prefix encoding used in our proposed \gls{tl}-based \gls{ppm} technique on the prediction performance compared to alternative prefix encoding approaches.
The two alternative prefix encoding approaches used in this evaluation were last-3-state encoding and aggregation encoding. Both are common prefix encoding approaches in outcome-oriented \gls{ppm}~\citep{Teinemaa.2019}.
The first approach, last-3-state encoding, encodes each prefix only by the last three events, or fewer, if the complete trace includes fewer than three events. The second approach, aggregation encoding, encodes each prefix in an aggregated form by calculating the mean value for each attribute across all events in the prefix.

We applied the index-based encoding, the last-3-state encoding, and the aggregation encoding to our proposed \gls{tl}-based \gls{ppm} technique for the inter-organizational use case. We also applied these three encoding approaches to two traditional \gls{ppm} techniques, using \gls{lstm} and \gls{xgb} models. 
We selected these two traditional \gls{ppm} techniques because the first uses the same \gls{lstm} model as in our proposed technique but not in an \gls{tl}-setting, and the second uses an \gls{xgb} model, which performed relatively well in the first evaluation of our inter-case organizational use case (see~Section~\ref{sec:results:inter_trad}).

Table~\ref{tab:ResultsSequenceEncoding} shows that our proposed technique with the index-based prefix encoding clearly outperforms all baseline approaches in terms of prediction performance on the test event data of the target context. This suggests that index-based encoding is an effective prefix encoding approach for our proposed \gls{tl}-based \gls{ppm} technique.

\newpage
\begin{table}[ht!]
\centering
\caption{Prediction performance of our proposed \gls{tl}-based \gls{ppm} technique and traditional \gls{ppm} techniques with different prefix encoding approaches for the inter-organizational use case (average and standard deviation over five runs).}
\label{tab:ResultsSequenceEncoding}
\resizebox{\textwidth}{!}{ 
\begin{tabular}{@{}lrrrrr@{}}
\toprule
\textbf{}                                       \textbf{ML approach}      & \textbf{Precision} & \textbf{Recall} & \textbf{F1-score} & \textbf{MCC} &\textbf{$\text{AUC}_{\text{ROC}}$} 
\\ \midrule

\multicolumn{6}{l}{\textsc{LSTM Model with Activity Embedding and Relative Cross-domain Timestamp Encoding}}\\
\textbf{Index-based encoding}                       &                   &                   &                   &              &    \\
Train and test on $L_S$    & 0.735 ($\pm$.003)   & 0.694 ($\pm$.003)     &  0.708 ($\pm$.002)  &  0.293 ($\pm$.007) &  0.730 ($\pm$.005)\\
Train and test on $L_T$    & 0.769 ($\pm$.012) &  0.769 ($\pm$.037)     &  0.766 ($\pm$.024)  & 0.330 ($\pm$.034) & 0.775 ($\pm$.018)\\
Train on $L_S$, test on $L_T$ & 0.739 ($\pm$.020)&  0.755 ($\pm$.050) &   \textbf{0.724} ($\pm$.021) &   0.206 ($\pm$.029) & \textbf{0.711} ($\pm$.020) \\

\textbf{Last-3-state encoding}                       &                   &                   &                   &              &    \\
Train and test on $L_S$    & 0.745 ($\pm$.003)   & 0.680 ($\pm$.009)     &  0.698 ($\pm$.007)  &  0.310 ($\pm$.003)&  0.745 ($\pm$.001)\\
Train and test on $L_T$           & 0.778 ($\pm$.020) &  0.792 ($\pm$.026)     &  0.779 ($\pm$.017)        & 0.351 ($\pm$.044) & 0.779 ($\pm$.013)\\
Train on $L_S$, test on $L_T$    & 0.725 ($\pm$.014)&  0.694 ($\pm$.054) &   0.699 ($\pm$.028) &   0.195 ($\pm$.039)       & 0.709 ($\pm$.025) \\

\textbf{Aggregation encoding}                       &                   &                   &                   &              &    \\
Train and test on $L_S$    & 0.724 ($\pm$.016)   & 0.697 ($\pm$.015)     &  0.704 ($\pm$.003)  &  0.265 ($\pm$.039) &  0.708 ($\pm$.025)\\
Train and test on $L_T$    & 0.731 ($\pm$.004)   & 0.741 ($\pm$.018)     &  0.734 ($\pm$.008)  &  0.225 ($\pm$.015) &  0.706 ($\pm$.013)\\
Train on $L_S$, test on $L_T$ & 0.718 ($\pm$.023) & \textbf{0.756} ($\pm$.046) &  0.685 ($\pm$.007)  &  0.074 ($\pm$.015) &  0.542 ($\pm$.027)\\
 
\midrule

\multicolumn{6}{l}{\textsc{LSTM Model with One-hot Activity Encoding and Min-max time Feature Encoding}}\\
\textbf{Index-based encoding}                       &                   &                   &                   &              &    \\
Train and test on $L_S$    & 0.737  ($\pm$.002)    &   0.677  ($\pm$.004)  &  0.695  ($\pm$.003) & 0.292  ($\pm$.005) & 0.721  ($\pm$.005) \\
Train and test on $L_T$           &  0.676  ($\pm$.047) & 0.775  ($\pm$.000)  & 0.678  ($\pm$.001) & 0.011  ($\pm$.016) & 0.567  ($\pm$.002) \\
Train on $L_S$, test on $L_T$    &    0.050 ($\pm$.000)       &  0.224  ($\pm$.000)         &   0.082  ($\pm$.000)        &   -0.029 ($\pm$.000)  & 0.430 ($\pm$.009) \\

\textbf{Last-3-state encoding}                       &                   &                   &                   &              &    \\
Train and test on $L_S$    & 0.745 ($\pm$.002)   & 0.675 ($\pm$.005)     & 0.694 ($\pm$.004)  & 0.307 ($\pm$.002) & 0.741 ($\pm$.002)\\
Train and test on $L_T$    & 0.624 ($\pm$.050)   & 0.776 ($\pm$.000)     & 0.678 ($\pm$.000)  & 0.004 ($\pm$.009) & 0.556 ($\pm$.004)\\
Train on $L_S$, test on $L_T$ & 0.143 ($\pm$.208) & 0.225 ($\pm$.001)     & 0.084 ($\pm$.003)  & -0.007 ($\pm$.015) & 0.442 ($\pm$.014)\\

\textbf{Aggregation encoding}                       &                   &                   &                   &              &    \\
Train and test on $L_S$    & 0.725 ($\pm$.002)   & 0.702 ($\pm$.005)     & 0.711 ($\pm$.003)  & 0.271 ($\pm$.004) & 0.713 ($\pm$.003)\\
Train and test on $L_T$    & 0.602 ($\pm$.000)   & 0.776 ($\pm$.000)     & 0.678 ($\pm$.000)  & 0.000 ($\pm$.000)   & 0.528 ($\pm$.003)\\
Train on $L_S$, test on $L_T$ & 0.050 ($\pm$.000)  & 0.224 ($\pm$.000)     & 0.082 ($\pm$.000)  & -0.029 ($\pm$.000)& 0.450 ($\pm$.027)\\
 
\midrule

\multicolumn{6}{l}{\textsc{XGBoost with One-hot Activity Encoding}}\\
\textbf{Index-based encoding}                       &                   &                   &                   &              &    \\
Train and test on $L_S$    & 0.714 ($\pm$.002)   & 0.743 ($\pm$.002)     & 0.720 ($\pm$.001)  & 0.234 ($\pm$.004) & 0.738 ($\pm$.002)\\
Train and test on $L_T$    & 0.793 ($\pm$.000)   & 0.811 ($\pm$.000)     & 0.793 ($\pm$.000)  & 0.386 ($\pm$.000) & 0.759 ($\pm$.000)\\
Train on $L_S$, test on $L_T$ & 0.828 ($\pm$.003) & 0.257 ($\pm$.044)     & 0.143 ($\pm$.080)  & 0.086 ($\pm$.059) & 0.604 ($\pm$.004)\\

\textbf{Last-3-state encoding}                       &                   &                   &                   &              &    \\
Train and test on $L_S$    & 0.729 ($\pm$.000)   & 0.750 ($\pm$.000)     & 0.735 ($\pm$.000)  & 0.279 ($\pm$.001) & 0.759 ($\pm$.000)\\
Train and test on $L_T$    & 0.782 ($\pm$.002)   & 0.801 ($\pm$.002)    & 0.785 ($\pm$.002)  & 0.361 ($\pm$.006) & 0.763 ($\pm$.002)\\
Train on $L_S$, test on $L_T$ & 0.707 ($\pm$.003) & 0.541 ($\pm$.002)    & 0.578 ($\pm$.002)  & 0.128 ($\pm$.005) & 0.616 ($\pm$.002)\\

\textbf{Aggregation encoding}                       &                   &                   &                   &              &    \\
Train and test on $L_S$    & 0.723 ($\pm$.000)   & 0.743 ($\pm$.001)     & 0.730 ($\pm$.000)  & 0.265 ($\pm$.001) & 0.749 ($\pm$.001)\\
Train and test on $L_T$    & 0.736 ($\pm$.002)   & 0.759 ($\pm$.004)     & 0.744 ($\pm$.002) & 0.236 ($\pm$.004) & 0.692 ($\pm$.003)\\
Train on $L_S$, test on $L_T$ & \textbf{0.831} ($\pm$.000) & 0.378 ($\pm$.011)     & 0.351 ($\pm$.016) & \textbf{0.225} ($\pm$.009) & 0.614 ($\pm$.002)\\

\bottomrule
\multicolumn{6}{p{18cm}}{\textit{Note.} Best results for training on $L_S$ and testing on $L_T$ are marked in bold.} 
\end{tabular}
}
\end{table}

\newpage
\section{Prediction performance for encoding activity and timestamp information of prefixes separately compared to encoding both jointly}
\label{app:PrefixIsoJointPrefixEncoding}

We conducted an additional evaluation to test whether traditional, text-based contextualized embedding models can effectively encode activity and timestamp information about prefixes in a joint manner. 
For this evaluation, we constructed a baseline technique that jointly encodes the activity and timestamp information about prefixes using one of the four pre-trained contextualized embedding models, which we introduced in the setup of our evaluation. Specifically, the information about a prefix of the length $N$ is fed into the embedding models as \textquote{$a_1$; $t_1$; $\dots$; $a_N$; $t_N$}, where $a$ is an activity and $t$ is a timestamp in the format \%d-\%m-\%Y \%H:\%M:\%S. 

The encoded prefixes are then mapped via an \gls{mlp} model onto the values of the \textit{in-time} prediction target. The \gls{mlp} model has three fully connected blocks, each of which is followed by tanh activation, batch normalization, and dropout. In addition, the model incorporates residual connections after the second and third blocks using projection layers. The final output is passed through a sigmoid activation function. As with all other \gls{dl} models used in our evaluation, the internal model parameters were optimized using the \gls{adam} optimizer \citep{kingma2015adam}.

Table~\ref{tab:ResultsTraceEncoding} shows that our \gls{tl}-based \gls{ppm} technique achieves a clearly higher prediction performance when trained on $L_S$ and tested on $L_T$ compared to the baseline techniques using the four different pre-trained contextualized embedding models. This suggests that it is ineffective to use traditional, text-based, contextualized embedding models to jointly encode activity and timestamp information about prefixes in our \gls{tl} setting.

\begin{table}[ht!]
\centering
\caption{Prediction performance for our \gls{tl}-based \gls{ppm} technique that encodes activity and timestamp information about a trace separately compared to a baseline technique that encodes activity and timestamp information about a trace jointly. The baseline technique was tested using the four pre-trained contextualized embedding models of our setup for the inter-organizational use case (average and standard deviation over five runs).}
\label{tab:ResultsTraceEncoding}
\resizebox{\textwidth}{!}{ 
\begin{tabular}{@{}lrrrrr@{}}
\toprule
\textbf{}                                        \textbf{Embedding model}     & \textbf{Precision} & \textbf{Recall} & \textbf{F1-score} & \textbf{MCC} &\textbf{$\text{AUC}_{\text{ROC}}$}
\\ \midrule

\multicolumn{6}{l}{\textsc{Separate Encoding of Activity and Timestamp Information}}\\

\textbf{Proposed approach}                      &                   &                   &                   &              &    \\
Train and test on $L_S$    & 0.735 ($\pm$.003)   & 0.694 ($\pm$.003)     &  0.708 ($\pm$.002)  &  0.293 ($\pm$.007) &  0.730 ($\pm$.005)\\
Train and test on $L_T$    & 0.769 ($\pm$.012) &  0.769 ($\pm$.037)     &  0.766 ($\pm$.024)  & 0.330 ($\pm$.034) & 0.775 ($\pm$.018)\\
Train on $L_S$, test on $L_T$ & \textbf{0.739} ($\pm$.020)&  \textbf{0.755} ($\pm$.050) &   \textbf{0.724} ($\pm$.021) &   \textbf{0.206} ($\pm$.029) & \textbf{0.711} ($\pm$.020) \\
\midrule
\multicolumn{6}{l}{\textsc{Joined Encoding of Activity and Timestamp Information}}\\

\textbf{bert-base-cased}                      &                   &                   &                   &              &    \\
Train and test on $L_S$           & 0.725 ($\pm$.006) & 0.591 ($\pm$.031) & 0.616 ($\pm$.031) & 0.233 ($\pm$.008) & 0.680 ($\pm$.005) \\
Train and test on $L_T$           & 0.602 ($\pm$.000) & 0.776 ($\pm$.000) & 0.678 ($\pm$.000) & 0.000 ($\pm$.000) & 0.517 ($\pm$.020) \\
Train on $L_S$, test on $L_T$     & 0.688 ($\pm$.006) & 0.466 ($\pm$.035) & 0.499 ($\pm$.038) & 0.075 ($\pm$.016) & 0.566 ($\pm$.007) \\

\textbf{bert-base-uncased}                      &                   &                   &                   &              &    \\
Train and test on $L_S$    & 0.714 ($\pm$.008)   & 0.666 ($\pm$.010)     &  0.682 ($\pm$.006)  &  0.238 ($\pm$.016)&  0.682 ($\pm$.006)\\
Train and test on $L_T$           & 0.602 ($\pm$.000) &  0.776 ($\pm$.000)     &  0.678 ($\pm$.000)        & 0.000 ($\pm$.000) & 0.526 ($\pm$.013)\\
Train on $L_S$, test on $L_T$    & 0.727 ($\pm$.025)&  0.397 ($\pm$.069) &   0.393 ($\pm$.101) &   0.109 ($\pm$.015)       & 0.569 ($\pm$.004) \\

\textbf{all-MiniLM-L12-v2}                      &                   &                   &                   &              &    \\
Train and test on $L_S$    & 0.649 ($\pm$.086)   & 0.668 ($\pm$.078)     & 0.638 ($\pm$.029)  & 0.131 ($\pm$.120) & 0.598 ($\pm$.096)\\
Train and test on $L_T$    & 0.620 ($\pm$.041)    & 0.750 ($\pm$.057)      & 0.676 ($\pm$.005)  & 0.023 ($\pm$.052) & 0.537 ($\pm$.055)\\
Train on $L_S$, test on $L_T$ & 0.645 ($\pm$.040) & 0.713 ($\pm$.076)     & 0.670 ($\pm$.028)   & 0.036 ($\pm$.035) & 0.540 ($\pm$.044)\\

\textbf{all-mpnet-base-v2}                      &                   &                   &                   &              &    \\
Train and test on $L_S$           & 0.651 ($\pm$.087)   & 0.688 ($\pm$.054)     &  0.657 ($\pm$.021)  &  0.140 ($\pm$.131) &  0.609 ($\pm$.102)\\
Train and test on $L_T$           & 0.656 ($\pm$.031)   & 0.763 ($\pm$.014)     &  0.684 ($\pm$.006)  &  0.021 ($\pm$.021) &  0.594 ($\pm$.009)\\
Train on $L_S$, test on $L_T$     & 0.636 ($\pm$.033)   & 0.772 ($\pm$.003)     &  0.679 ($\pm$.002)  &  0.006 ($\pm$.012) &  0.527 ($\pm$.036)\\

\bottomrule
\multicolumn{6}{p{18cm}}{\textit{Note.} Best results for training on $L_S$ and testing on $L_T$ are marked in bold.}
\end{tabular}
}
\end{table}

\end{document}